%% file: mainv4.tex
\documentclass[10pt,journal,compsoc]{IEEEtran}
\usepackage[ruled,linesnumbered]{algorithm2e}
\usepackage{multirow}
\usepackage[table]{xcolor}
\usepackage{amssymb}
\usepackage{amsfonts}
\usepackage{multicol}
\usepackage{graphicx}
\usepackage{booktabs}
\usepackage[T1]{fontenc}
\usepackage[breaklinks=true,bookmarks=true,pagebackref=false,colorlinks,linkcolor=blue,anchorcolor=red,citecolor=blue,urlcolor=blue]{hyperref}
\usepackage[numbers,sort]{natbib}

\usepackage{comment}
\usepackage{subcaption}
\usepackage{amsmath}
\usepackage{algorithmic}
\usepackage{array}
\usepackage[switch]{lineno}
\usepackage{longtable}
\usepackage{xspace}
\usepackage{url}
\usepackage{overpic}
\usepackage{ragged2e}
\usepackage{framed}
\usepackage{enumitem}
\usepackage{balance}
\input{shortcuts}

\begin{document}
%
\title{DIVER: Reinforced Diffusion Breaks Imitation Bottlenecks in End-to-End Autonomous Driving}
\author{Ziying Song, Lin Liu, Hongyu Pan,  Bencheng Liao, Mingzhe Guo, Lei Yang, \\ Yongchang Zhang,   Shaoqing Xu, Caiyan Jia, Yadan Luo
\thanks{Corresponding author: Caiyan Jia.}
\IEEEcompsocitemizethanks{
\IEEEcompsocthanksitem Ziying Song is with the Beijing Key Laboratory of Traffic Data Mining and Embodied Intelligence, School of Computer Science and Technology, Beijing Jiaotong University, Beijing, China, and also with the School of Artificial Intelligence (School of Software), Yanshan University, Qinhuangdao, China. E-mail: 22110110@bjtu.edu.cn and songziying@ysu.edu.cn. Ziying Song contributed to this work in part while serving as an intern at Horizon Robotics.
\IEEEcompsocthanksitem Lin Liu, Mingzhe Guo and Caiyan Jia are with Beijing Key Laboratory of Traffic Data Mining and Embodied Intelligence, School of Computer Science and Technology, Beijing Jiaotong University.
\IEEEcompsocthanksitem Hongyu Pan, Bencheng Liao, and Yongchang Zhang  are with Horizon Robotics.
\IEEEcompsocthanksitem Lei Yang is with School of Mechanical and Aerospace Engineering, Nanyang Technological University, Singapore.
\IEEEcompsocthanksitem Shaoqing Xu is with  University of Macau, China 
\IEEEcompsocthanksitem Yadan Luo is with the School of Electrical Engineering and Computer Science, The University of Queensland, Australia. 
}


}

\markboth{IEEE Transactions on Pattern Analysis and Machine Intelligence
}%
{Song \MakeLowercase{\textit{et al.}}: 
DIVER: Reinforced Diffusion Breaks Imitation Bottlenecks in End-to-End Autonomous Driving
}
%
%
\IEEEtitleabstractindextext{%
\begin{abstract}
\justifying 
Existing end-to-end autonomous driving (E2E-AD) methods predominantly rely on single expert demonstration through imitation learning, often leading to conservative and homogeneous driving behaviors that struggle to generalize to complex real-world scenarios. In this work, we propose \textbf{DIVER}, a novel E2E-AD framework that combines diffusion-based multi-mode trajectory generation with reinforcement learning to produce diverse, safe, and goal-directed trajectories. First, the model conditions on map elements and surrounding agents to generate multiple reference trajectories from each ground-truth reference trajectory that overcome the inherent limitations of single-mode imitation. Second, we treat the diffusion process as a stochastic policy and employ Group Relative Policy Optimization (GPRO) objectives to guide the diffusion process. By optimizing trajectory-level rewards for both diversity and safety, GRPO directly mitigates mode collapse and enhances collision avoidance, encouraging exploration beyond expert demonstrations and ensuring physically plausible plans. Furthermore, to address the limitations of L2-based open-loop metrics in capturing trajectory diversity, we propose a novel trajectory diversity metric to evaluate the diversity of multi-mode predictions. Extensive experiments on the closed-loop NAVSIM and Bench2Drive benchmarks, as well as the open-loop nuScenes dataset, demonstrate that DIVER significantly improves trajectory diversity, effectively addressing the mode collapse problem inherent in imitation learning. The source code is available at \url{https://github.com/adept-thu/diver}.
\end{abstract}

\begin{IEEEkeywords}
Autonomous Driving, End-to-End Autonomous Driving, Diffusion Model, Reinforcement Learning.
\end{IEEEkeywords}}

\maketitle

\IEEEdisplaynontitleabstractindextext

\IEEEpeerreviewmaketitle

\IEEEraisesectionheading{
\section{Introduction}\label{sec:introduction}
}
\begin{figure}[t]
\centering
 \includegraphics[width=1.0\linewidth]{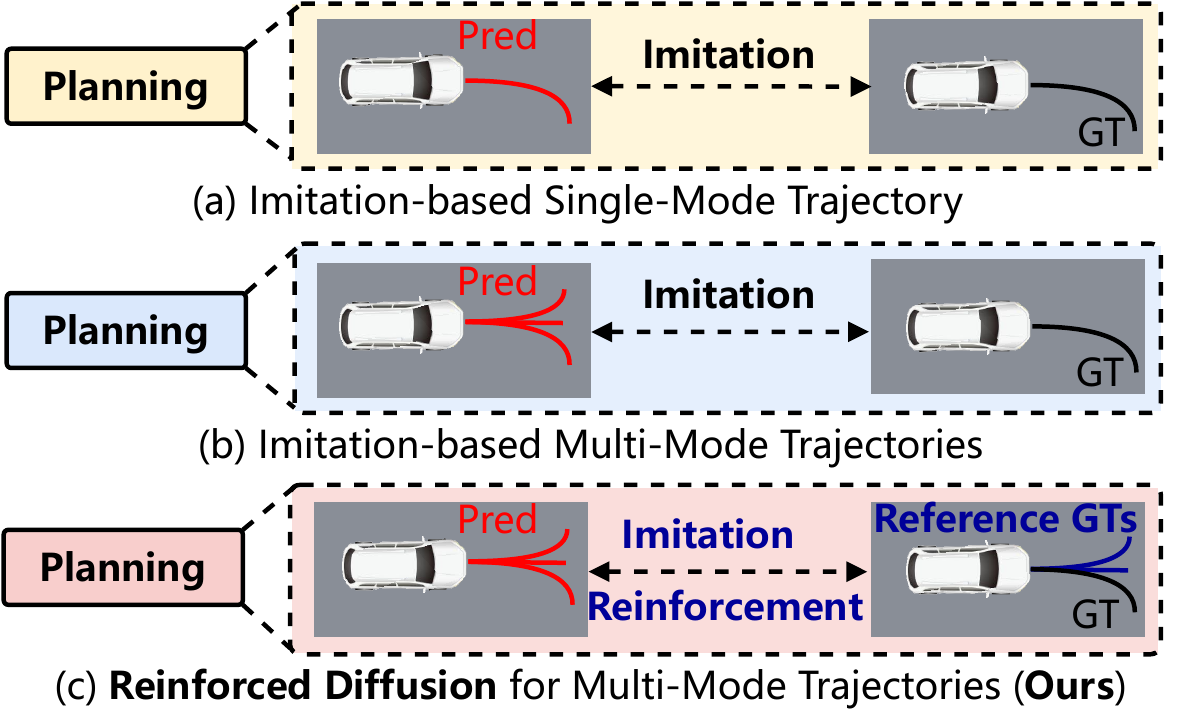}
\caption[ ]{\textcolor{black}{\textbf{(a) Imitation-based Single-Mode Trajectory Planning} \cite{uniad, jiang2023vad,  TransFuser} predicts deterministic trajectories but lacks action diversity, leading to potential safety risks. 
\textbf{(b) Imitation-based Multi-Mode Trajectory  Planning} \cite{sun2024sparsedrive, chen2024vadv2, liao2024diffusiondrive} fails to address the diversity loss in  imitation learning end-to-end autonomous driving, leading to mode collapse. The generated multi-mode trajectories overly depend on a single GT trajectory, ultimately clustering around it. 
\textbf{(c) The proposed DIVER framework} adopts reinforced diffusion for multi-mode trajectory generation, encouraging the ego-vehicle to produce diverse driving behaviors instead of rigidly following a single expert.
}}
\label{fig:motivation}
\end{figure}
\IEEEPARstart{I}{n} recent years, autonomous driving has entered a new phase, with significant progress in end-to-end autonomous driving (E2E-AD)  \cite{LiHongyange2etpamisurvey, song2024robustness}. E2E-AD systems integrate traditionally independent tasks, such as 3D object detection, multi-object tracking, online mapping, motion prediction, and planning, into a unified framework that directly learns driving policies from raw sensor inputs  \cite{uniad, jiang2023vad, sun2024sparsedrive, liao2024diffusiondrive, TransFuser}. Among these tasks, planning serves as a core module that not only directly governs ego-vehicle decision-making and control but also provides implicit guidance to upstream perception tasks  \cite{uniad}.

Currently, imitation learning (IL) remains the dominant paradigm for trajectory planning in E2E-AD systems \cite{LiHongyange2etpamisurvey,jia2025progressive}. IL enables E2E-AD models to mimic human expert driving behaviors through supervised learning, typically by minimizing the discrepancies between predicted and ground-truth (GT) trajectories. The early E2E-AD methods  \cite{uniad, jiang2023vad,liu2025fully,liu2026driveworld,TransFuser} mainly adopted single-mode trajectory generation, as shown in Figure \ref{fig:motivation}(a), where a single deterministic trajectory is regressed toward the expert demonstration. However, such single-mode designs struggle in complex traffic scenarios, as they fail to capture environmental uncertainty and interactions among dynamic agents.
To mitigate these limitations, subsequent studies  \cite{chen2024vadv2,sun2024sparsedrive,liao2024diffusiondrive,li2024hydra} have proposed \textit{multi-mode} trajectory prediction frameworks, as illustrated in Figure \ref{fig:motivation}(b). For instance, VADv2  \cite{chen2024vadv2} employs probabilistic modeling to infer potential driving intentions, generating multiple trajectory proposals and selecting the optimal one via a scoring mechanism. However, these approaches  \cite{chen2024vadv2,sun2024sparsedrive,liao2024diffusiondrive,li2024hydra} still suffer from the issue of \textit{mode collapse}, where multi-mode trajectories remain overly concentrated around the GT path, reflecting excessive dependence on expert demonstrations and limited behavioral diversity. This diversity \textit{bottleneck} stems from the inherent nature of IL, as models are trained to imitate a single expert trajectory in both open-loop and closed-loop settings. As a result, their ability to explore and represent diverse driving behaviors remains fundamentally constrained.

To enhance diversity, several E2E works  \cite{liao2024diffusiondrive,guo2025vdt,BridgeDrive,EnDfuser} have recently introduced diffusion models to generate diverse trajectories. Diffusion models  \cite{ddpm}, as a powerful generative learning framework, have demonstrated strong capabilities in modeling multi-mode behavior distributions for robotic policy learning. Introducing diffusion models into E2E-AD trajectory planning has emerged as a promising research direction.
\textcolor{black}{For instance, DiffusionDrive \cite{liao2024diffusiondrive} integrates truncated diffusion strategies with efficient cascaded diffusion decoders. VDT-Auto \cite{guo2025vdt} geometrically and contextually parses the environment to condition the diffusion process, while EnDfuser \cite{EnDfuser} generates a distribution of candidate trajectories from a single perception frame through ensemble diffusion. BridgeDrive \cite{BridgeDrive} further introduces a principled diffusion framework that translates coarse anchors into fine-grained trajectory plans.
However, as their optimization processes still adhere to imitation learning (IL) frameworks based on single expert trajectories, the generated multi-mode trajectories ultimately tend to converge toward a single ground truth. This limitation prevents these models from fully overcoming the inherent constraints of imitation learning.}

In this work, we propose \textbf{DIVER},  a novel multi-mode E2E-AD framework that leverages reinforcement learning to guide diffusion models for generating diverse, feasible trajectories as shown in Figure \ref{fig:motivation}(c).
By leveraging diffusion models, our approach generates multiple plausible future trajectories, capturing a spectrum of driving behaviors while maintaining safety and feasibility.
Specifically, we introduce the Policy-Aware Diffusion Generator (PADG), which incorporates map elements and surrounding agents as conditions and generates diverse multi-mode trajectories and multiple reference GTs, guided by a single GT trajectory and trajectory anchors, effectively capturing various driving styles. To ensure both safety and diversity in trajectory planning, we incorporate reinforcement learning to supervise the diffusion process via reward signals, enabling optimization of generated trajectories under diversity and safety constraints. Additionally, to address the misalignment between conventional open-loop metrics (primarily L1 Loss) and the intrinsic requirement for diverse trajectory generation, we introduce a novel Diversity Metric to better evaluate the diversity of multi-mode trajectory predictions.
Extensive experiments on closed-loop \textbf{Bench2Drive} and \textbf{NAVSIM}, as well as open-loop \textbf{nuScenes}, demonstrate that our DIVER significantly enhances trajectory diversity and planning robustness, effectively addressing the mode collapse issue in imitation learning.

Overall, our contributions are as follows:
\begin{itemize}
    \item  We propose the \textbf{DIVER}, an novel multi-mode E2E-AD framework that uses reinforcement learning to guide diffusion models in generating diverse and feasible driving behaviors.

    \item We introduce the \textbf{Policy-Aware Diffusion Generator (PADG)}, which incorporates map elements and agent interactions as conditional inputs, enabling the generation of multi-mode trajectories that capture diverse driving styles.

    \item  We leverage reinforcement learning to guide the diffusion model with diversity and safety rewards, addressing the limitations of imitation learning.

    \item We propose a novel \textbf{Diversity Metric} to evaluate multi-mode trajectory generation, providing a more principled way to assess the diversity and effectiveness of generated trajectories compared to existing metrics.
    
    \item Extensive evaluations on the Bench2Drive, NAVSIM, NuScenes demonstrate that DIVER significantly improves the diversity, safety, and feasibility of generated trajectories over state-of-the-art methods.
\end{itemize}

\section{Related Work}\label{sec:related_work}
\subsection{End-to-end Autonomous Driving}
End-to-end autonomous driving is a fully differentiable machine learning system that takes raw sensor input data and other metadata as prior information, directly outputting control signals or trajectory planning for vehicles  \cite{song2024robustness,LiHongyange2etpamisurvey}.
A series of E2E-AD works  \cite{uniad,jiang2023vad,TransFuser,liu2025guideflow,sun2025focalad,sun2025minddrive,hedrive_zhangxingyu}, represented by UniAD \cite{uniad} and VAD \cite{jiang2023vad}, have adopted single-mode trajectory for planning performance. However, single-mode trajectory planning methods have issues with action diversity and associated safety risks.
To address the above limitations, VADv2  \cite{chen2024vadv2} has recently shifted to a multi-mode planning framework by scoring and sampling from an extensive fixed vocabulary of anchor trajectories. Based on the design of multi-mode planning, SparseDrive  \cite{sun2024sparsedrive} proposes a hierarchical planning selection strategy to select a rational and safe trajectory as the final planning output.
By leveraging historical trajectory momentum, MomAD  \cite{momad} addresses the issue of inconsistent multi-mode trajectories over time.
Existing multi-mode trajectory planning methods  \cite{chen2024vadv2,sun2024sparsedrive,momad} do not fundamentally resolve mode collapse in imitation learning.
\textcolor{black}{
DriveSuprim  \cite{yao2025drivesuprim} advances selection-based multi-mode trajectory generation via coarse-to-fine candidate filtering, rotation-based augmentation, and self-distillation, while achieving state-of-the-art safety and trajectory quality. Hydra-MDP  \cite{li2024hydra} uses teacher-student knowledge distillation to incorporate human and rule-based guidance via metric distillation. Overall, our DIVER combines diffusion-based generative modeling with reinforcement learning, thereby bridging generative trajectory modeling and task-aware policy optimization.
}

\subsection{Diffusion Models for Autonomous Driving}

Diffusion models approximate data distributions through iterative denoising and have demonstrated remarkable performance in image generation tasks  \cite{yang2023diffusion}, showing great potential for behavior modeling and trajectory generation in autonomous driving \cite{wang2024drivedreamer}. DiffusionDrive  \cite{liao2024diffusiondrive} leverages diffusion models to capture multi-mode action distributions in an E2E-AD framework. DiffScene \cite{xu2025diffscene} utilizes diffusion models to synthesize safe scenarios for evaluating autonomous driving safety. DiffBEV  \cite{zou2024diffbev} adopts conditional diffusion to produce fine-grained BEV representations. MotionDiffuser  \cite{jiang2023motiondiffuser} employs diffusion-based representations for multi-agent trajectory prediction. DiffusionDrive  \cite{liao2024diffusiondrive} models multi-mode trajectory distributions via a truncated diffusion process. VDT-Auto  \cite{guo2025vdt} integrates diffusion Transformers with VLMs for action generation.  DiFSD  \cite{su2024difsd} introduces both position-level motion diffusion and trajectory-level planning denoising to model uncertainty, thereby enhancing the stability and convergence of the entire training framework. Overall, diffusion-based research in E2E-AD remains limited but holds great potential.

\subsection{Reinforcement Learning for Autonomous Driving}

Reinforcement Learning enables an agent to learn an optimal policy through interactions with the environment by maximizing cumulative long-term rewards  \cite{wang2022deep}. 
Landmark achievements such as AlphaGo  \cite{silver2016mastering} and AlphaGo Zero  \cite{silver2017mastering} have demonstrated RL’s remarkable capabilities in complex strategy games, 
while AlphaFold  \cite{jumper2021highly} has showcased its power in protein structure prediction. 
More recently, the successful application of RL in LLMs like DeepSeek-R1  \cite{guo2025deepseek} further validates its broad applicability and value.
In autonomous driving, several works have begun to explore the integration of RL. 
Toromanoff et al.  \cite{toromanoff2020end} introduced implicit affordances to enable RL to handle urban driving scenarios, including lane keeping, pedestrian and vehicle avoidance, and traffic light interpretation. 
Zhang et al.  \cite{zhang2021end} trained an RL expert to map bird’s-eye view images to low-level continuous actions, providing informative supervision for imitation learning agents. 
RAD  \cite{gao2025rad} trains an end-to-end driving agent in a photorealistic 3DGS environment using RL. 
AlphaDrive  \cite{jiang2025alphadrive} is among the first to integrate GRPO-based RL with planning and reasoning, substantially improving both performance and training efficiency in E2E-AD.

\section{Preliminary}
\subsection{End-to-End Autonomous Driving}\label{sec:e2e}
Current E2E-AD systems \cite{uniad, jiang2023vad, sun2024sparsedrive, liao2024diffusiondrive,  TransFuser} are predominantly based on IL, which aims to learn expert driving policies by mapping raw sensor inputs to ego-vehicle trajectory outputs. It consists of several sequential or unified tasks, including 3D object detection, multi-object tracking, online mapping, motion prediction, and planning. 3D object detection and multi-object tracking tasks provide dynamic semantic features of surrounding agents for planning, while the online mapping module extracts high-confidence lane and road elements as static map features to assist trajectory planning. Among these, planning plays a central role, as it determines the final decision of the ego vehicle.

\noindent \textbf{Multi-Mode Trajectories.}
The planning module typically predicts a sequence of future waypoints, represented as $\tau = \{(x_t, y_t)\}_{t=1}^{T_f}$, where $(x_t, y_t)$ denotes the ego-vehicle position in its local coordinate frame at time step $t$, and $T_f$ is the planning horizon. For datasets such as Bench2Drive \cite{jia2024bench2drive} and nuScenes \cite{nuscenes}, trajectories are usually sampled at 2 Hz over a 3-second horizon, resulting in 6 waypoints.
To enhance robustness, safety, and adaptability, E2E-AD systems increasingly incorporate diverse driving strategies. A common scheme to enable such diversity is through multi-mode trajectory prediction, where the system predicts $M$ future trajectories, typically $M = 6$, denoted as 
$\tau = \left\{ \left( x_t^{(m)}, y_t^{(m)}\right) \,\middle|\, t = 1, \dots, T_f;\; m = 1, \dots, M \right\}$. The final trajectory is selected from these candidates based on scoring functions, and the diversity among them (\textit{e.g.}, turning, overtaking, following) is used to infer the ego-vehicle’s potential driving strategies.

\subsection{Diffusion Models for Driving Policy Diversity}   
\begin{figure}[t]
\centering
 \includegraphics[width=0.5\linewidth]{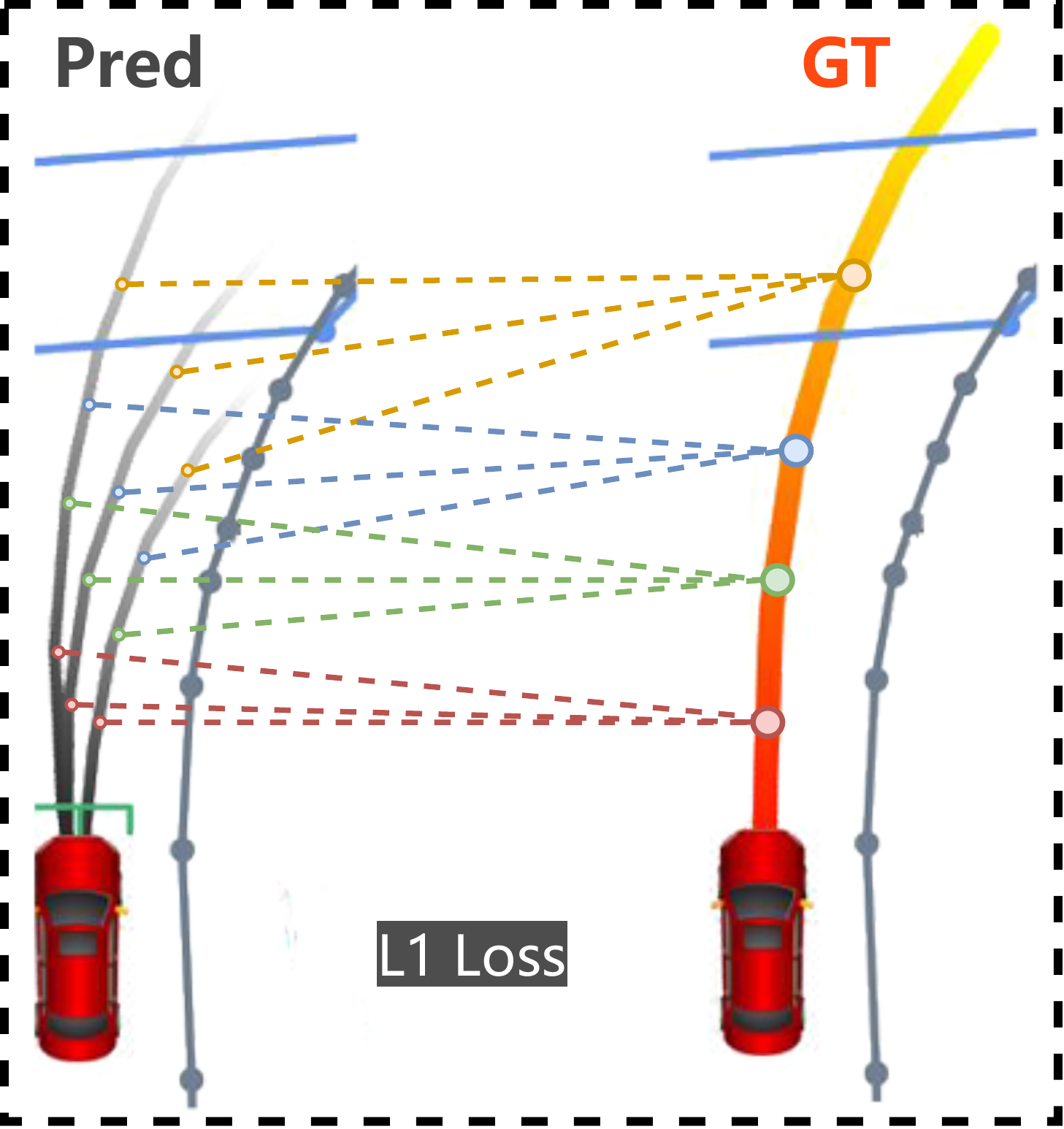}
\caption[ ]{
\textbf{Imitation learning-based multi-mode trajectory paradigm.}
Most IL-based multi-mode E2E-AD methods rely on L1 loss for training and L2 distance for evaluation, which emphasizes matching a single GT trajectory rather than modeling diversity. This misalignment limits the generation of truly diverse behaviors. Even with diffusion-based frameworks  \cite{liao2024diffusiondrive}, such imitation-driven objectives constrain their capacity to capture multi-mode driving patterns.
}
\label{fig:l1loss}
\end{figure}

Diffusion models, owing to their probabilistic generative nature and progressive denoising mechanism, offer unique advantages in modeling diverse driving strategies. Through iterative denoising, they can generate a wide distribution of high-quality candidate trajectories (\textit{e.g.}, lane changes, deceleration, or detours). 
However, most existing diffusion-based planning methods are still trained under the imitation learning (IL) paradigm, relying on a single expert trajectory as supervision, as shown in Figure \ref{fig:l1loss}. Under this setup, despite the strong generative capacity of diffusion models, the output multi-mode trajectories tend to collapse around the GT, exhibiting limited diversity. This raises a fundamental question: \textbf{Why do diffusion models struggle to generate diverse trajectories under imitation learning supervision?}

Under conventional imitation learning (IL) settings, where training is supervised by a single ground-truth (GT) trajectory, their ability to generate diverse behaviors is fundamentally constrained. We present a theoretical analysis to explain this limitation.  Let a trajectory be defined as a sequence of waypoints $\tau = {(x_t, y_t)}_{t=1}^{T_f}$, and a multi-mode output as $\{\tau^{(m)}\}_{m=1}^{M}$. In IL-based trajectory prediction, the training objective typically maximizes the log-likelihood of the observed trajectory:
\begin{equation}
\max_\theta \mathbb{E}_{\tau \sim p_{\text{data}}}\left[ \log p_\theta(\tau) \right].
\end{equation}
When the dataset provides only a single GT trajectory $\tau^*$ per input, the data distribution reduces to a Dirac delta:
\begin{equation}
p_{\text{data}}(\tau) = \delta(\tau - \tau^*).
\end{equation}
Consequently, the training objective simplifies to maximizing the likelihood of this unique trajectory:
\begin{equation}
\max_\theta \log p_\theta(\tau^*).
\end{equation}
This forces the model to concentrate all probability mass on $\tau^*$, yielding a degenerate solution:
\begin{equation}
p_\theta^{\text{opt}}(\tau) = \delta(\tau - \tau^*).
\end{equation}
In contrast, diffusion models are trained by minimizing a denoising score-matching loss, often expressed as:
\begin{equation}
\min_\theta \mathbb{E}_{\epsilon \sim \mathcal{N}(0,I)}\left[ \left\| \epsilon - \epsilon_\theta(\tau^* + \sqrt{\alpha_t}\epsilon, t) \right\|^2 \right].
\end{equation}
This formulation teaches the model to reconstruct $\tau^*$ from noisy perturbations, effectively collapsing all modes toward the single GT trajectory. No matter how strong the underlying generative process is, the diversity of output trajectories $\{\tau^{(m)}\}_{m=1}^{M}$collapses around $\tau^*$, failing to reflect genuine multi-modality.

We further analyze the diversity loss via the covariance of the output distribution:
\begin{equation}
\Sigma = \mathbb{E}_{\tau \sim p_\theta}\left[ (\tau - \mu)(\tau^* - \mu)^T \right], \quad \mu = \mathbb{E}[\tau].
\end{equation}
With all $\tau$ approximating $\tau^*$, we have $\mu \approx \tau^*$ and $\Sigma \approx 0$, indicating near-zero diversity across predicted trajectories.
In summary, diffusion-based E2E-AD methods trained under single-GT supervision inherit the same mode collapse issue as standard IL, despite their generative capacity. To address this, we propose a hybrid IL-RL training framework, DIVER, that introduces diversity-aware rewards to encourage multi-mode, safe trajectories generation.


\begin{figure*}[t]
\centering
 \includegraphics[width=1.0\linewidth]{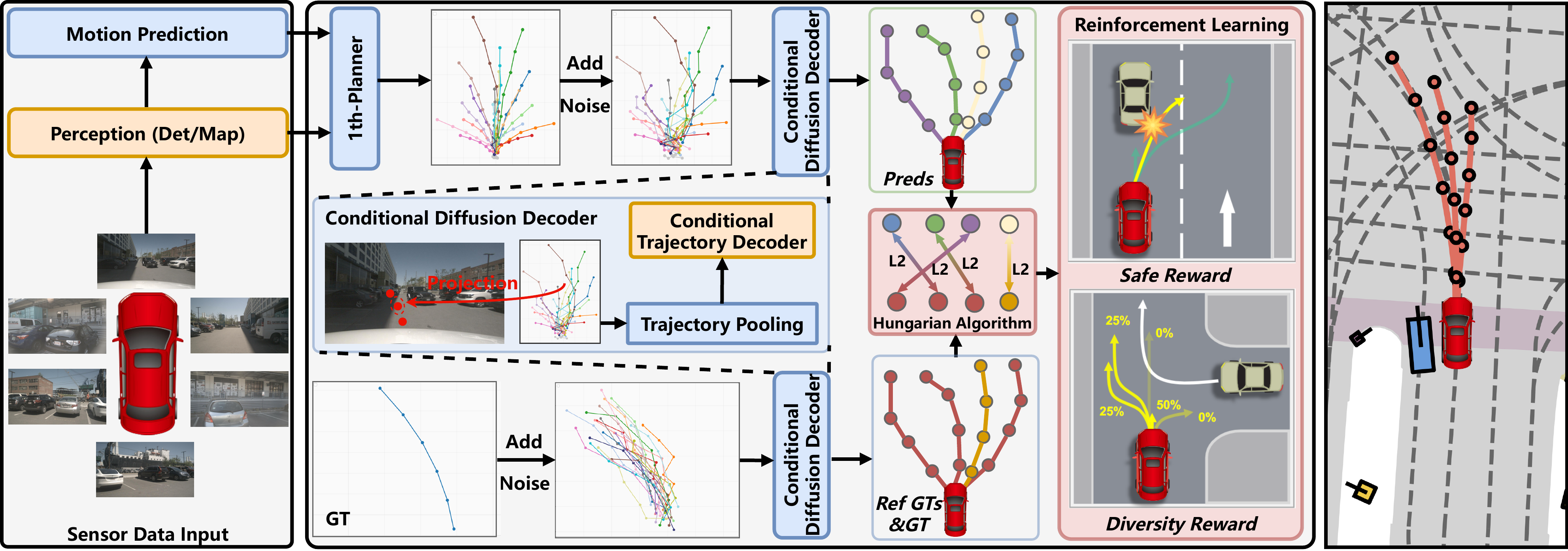}
\caption[ ]{\textbf{\textcolor{black}{The overall architecture of DIVER.}} As a multi-mode trajectory E2E-AD framework, DIVER first encodes multi-view images into feature maps to extract scene representations through a perception module. It then predicts the motion of surrounding agents and performs planning via a conditional diffusion model guided by reinforcement learning to generate diverse multi-mode trajectories. Our approach effectively addresses the inherent mode collapse in imitation learning, enabling the generation of safe and diverse behaviors for complex driving scenarios.

}
\label{fig:DIVER}
\end{figure*}

\section{DIVER}\label{sec:method}
\subsection{Overall Architecture}
\textcolor{black}{Existing E2E-AD methods  \cite{uniad, jiang2023vad, sun2024sparsedrive, liao2024diffusiondrive, TransFuser} predominantly rely on imitation learning (IL) from single expert demonstration, which often leads to conservative and homogeneous driving behaviors, thereby limiting their generalization to complex real-world scenarios. 
Although some diffusion-based methods, such as DiffusionDrive  \cite{liao2024diffusiondrive}, introduce diffusion models to mitigate mode collapse, their training remains constrained by imitation learning and lacks explicit optimization for task-level objectives such as safety or lane compliance.
}
Therefore, as shown in Figure \ref{fig:DIVER}, we propose \textbf{DIVER}, a novel E2E-AD framework that integrates diffusion models and reinforcement learning to enable diverse, and safe multi-mode trajectories. \textbf{DIVER} consists of two main components:  perception module and motion planner, which takes raw perception features as input and outputs a distribution over future trajectories. At the training paradigm level, \textbf{DIVER} adopts a hybrid learning scheme that combines IL with RL.
The core innovations of \textbf{DIVER} lie in two aspects. First, the \textbf{Policy-Aware Diffusion Generator (PADG)} generates diverse trajectory candidates by leveraging a diffusion model conditioned not only on a single expert trajectory but also on multiple Reference GTs, capturing a range of plausible human-like driving behaviors such as lane changes, yielding, and overtaking. Second, we leverage reinforcement learning to guide the diffusion-based trajectory generation process, addressing the inherent limitations of IL-based supervision and aligning the output with safety and diversity objectives.
Overall, \textbf{DIVER} combines the generative capability of diffusion models with the optimization strength of reinforcement learning, forming a flexible and scalable framework for E2E-AD. It enables the system to explore a broader solution space in complex scenarios while ensuring the physical feasibility and behavioral safety of the generated trajectories.

\subsection{Policy-Aware Diffusion Generator (PADG)}
To overcome the limitations of conservative and mode-collapsed behaviors in IL-based E2E-AD, we propose the PADG. PADG is a core component of our DIVER that enables diverse, and feasible trajectory generation through a conditional denoising diffusion process.
Specifically, PADG is built upon a conditional diffusion framework, which reconstructs multi-mode future trajectory distributions from random noise through iterative denoising, guided at each step by rich scene semantics as conditional information. It consists of a dual-branch architecture: one branch learns to reconstruct the distribution of the predicted trajectory, while the other models the expert trajectory to serve as guidance.

\subsubsection{Diverse Multi-Mode Trajectory Forward Diffusion}
To generate diverse and informative future trajectory distributions, we apply a forward diffusion process to multiple trajectory types, including multi-mode predicted trajectories and the GT trajectory. This step aims to perturb the trajectories with Gaussian noise in a controlled manner, enabling the conditional diffusion model to learn a robust denoising process that captures the underlying multi-modeity of future motion.

Specifically, let $\tau_0^{(m)}\in \mathbb{R}^{T \times 2}$ denote a predicted future trajectory of mode $m$, and $\tau_0^{\text{gt}}$ the GT future trajectory. At each diffusion step $t = 1, \dots, T$, the trajectories are corrupted with Gaussian noise following the standard forward diffusion process:
\begin{align}
q\left({\tau}^{(m)}_t \mid {\tau}^{(m)}_0\right) &= \mathcal{N}\left({\tau}^{(m)}_t; \sqrt{\alpha_t}{\tau}^{(m)}_0, (1 - \alpha_t)\mathbf{I}\right), \\
q\left(\tau^{\text{gt}}_t \mid \tau^{\text{gt}}_0\right) &= \mathcal{N}\left(\tau^{\text{gt}}_t; \sqrt{\alpha_t}\tau^{\text{gt}}_0, (1 - \alpha_t)\mathbf{I}\right),
\end{align}
where $\alpha_t \in (0, 1)$ is the noise schedule at timestep $t$. This formulation ensures that the model can learn to recover high-quality trajectory samples from various noise levels during training, while preserving the structured correlations of the predicted trajectories.

By jointly applying noise to predicted trajectories, we encourage the model to explore the latent trajectory space more broadly, leading to greater trajectory diversity during generation. Simultaneously, injecting noise into the GT trajectory supports supervised learning of the denoising objective, grounding the generative process in realistic motion patterns.

\begin{figure}[t]
\centering
 \includegraphics[width=1.0\linewidth]{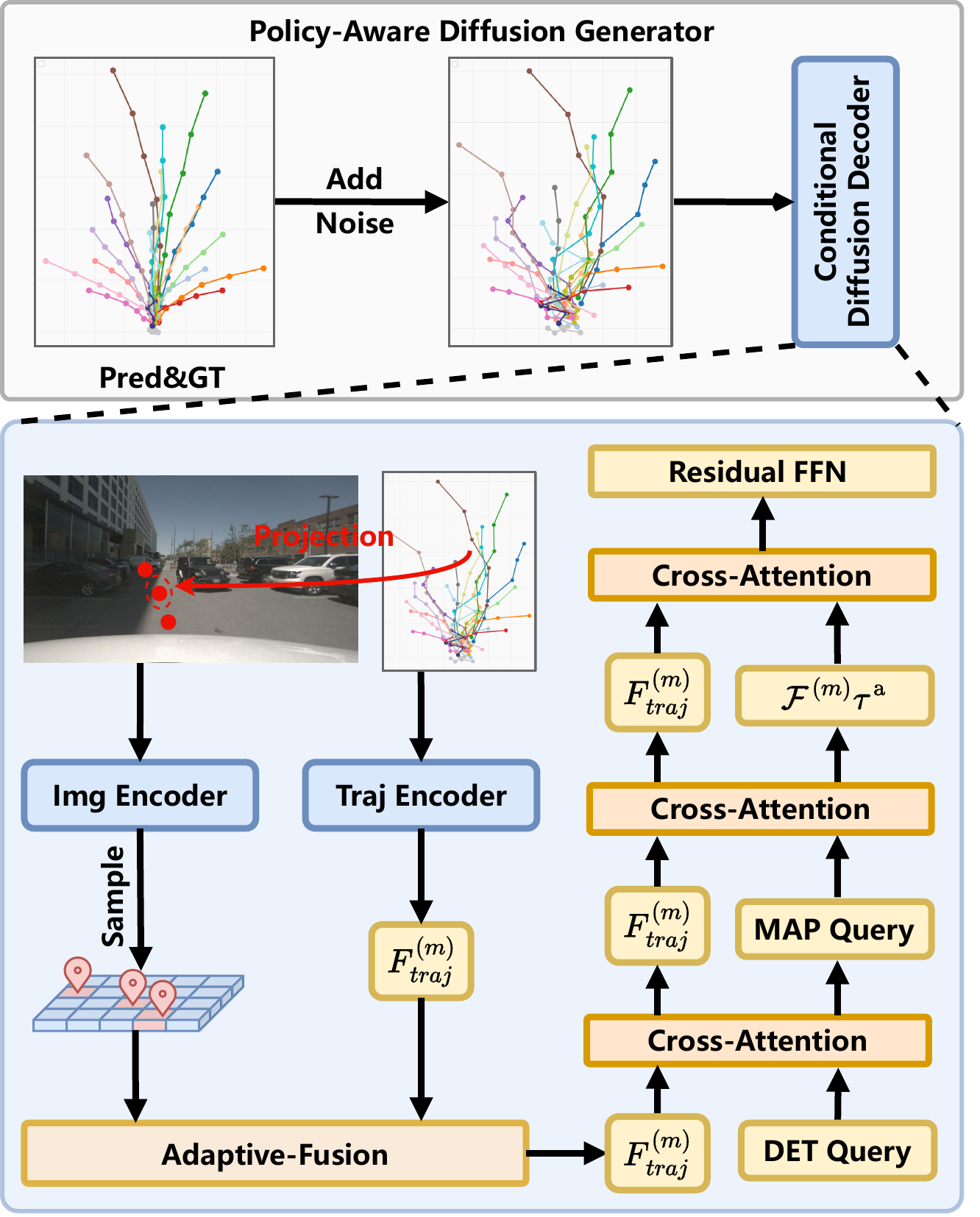}
\caption[ ]{\textbf{The illustration of PADG.} By incorporating the predicted trajectory, and GT trajectory as inputs, PADG reconstructs diverse multi-mode trajectories from noise through a conditional denoising process, guided by map and agent context.
}
\label{fig:PADG}
\end{figure}

\subsubsection{Conditional Diffusion Decoder} To effectively guide the denoising process in the conditional diffusion model, we incorporate rich scene semantics, including the BEV map, agent, and anchor trajectory features, into the trajectory generation process. 

Given noisy multi-mode trajectories $\tau_t^{(m)}$ and their corresponding anchor trajectories $\tau_t^{\text{a}(m)}$, we extract spatio-temporal features through sinusoidal position embedding and transformer-based encoding. Specifically, we first apply a forward diffusion process to perturb the trajectories with Gaussian noise. Let $\tilde{\tau}^{(m)}$ denote the noised multi-mode trajectories after de-normalization. We then construct high-dimensional embeddings as:
\begin{align}
\mathcal{E}^{(m)}_{\tau}&= \text{PE}(\tilde{\tau}^{(m)}), 
\mathcal{F}^{(m)}_{\tau}= \text{Enc}(\mathcal{E}^{(m)}_{\tau}), \\
\mathcal{E}^{(m)}_{\tau^{\text{a}}}&= \text{PE}(\tilde{\tau}^{\text{a}(m)}), 
\mathcal{F}^{(m)}_{\tau^{\text{a}}}= \text{Enc}(\mathcal{E}^{(m)}_{\tau^{\text{a}}}),
\end{align}
where $\text{PE}(\cdot)$ denotes sine-based positional encoding, and $\text{Enc}(\cdot)$ is a lightweight transformer encoder that captures the temporal dynamics and structural properties of the noisy trajectories. These embeddings are used for subsequent conditional interaction with the scene features.

To guide the generation of diverse and feasible future trajectories, we design a conditional diffusion decoder that performs multi-mode interaction between the noisy trajectory features and high-level semantic cues from the driving scene. This module also incorporates a GT-based reference trajectory to inject goal-oriented inductive bias.

Given the noisy GT trajectory $\tilde{\tau}^{\text{gt}}_t$ obtained from the forward diffusion process, we use it as the reference trajectory $\tau^{\text{ref}(m)}$ for feature embedding:
\begin{align}
\mathcal{F}_{\text{ref}}^{(m)}&= \text{PE}\left( \phi\left( \tau^{\text{ref}(m)}\right) \right),
\end{align}
where $\phi(\cdot)$ denotes sine-cosine positional embeddings applied to each trajectory waypoint, and $\text{PE}(\cdot)$ is a trajectory position encoder. Here, $\mathcal{F}_{\text{ref}}^{(m)}$ represents the learned feature of the noisy GT trajectory, which is then used for multi-mode interaction with high-level scene features in the conditional diffusion decoder.

To effectively model the interaction between noisy trajectory predictions and multi-mode contextual information, we design a two-stage conditional feature fusion pipeline, as detailed in Algorithm \ref{algorithm:TrajPooler} and \ref{algorithm:DiffusionTrajectoryDecoder}.

In the first stage, \textcolor{black}{we adopt a lightweight trajectory-aware feature aggregation module, termed TrajPooler (Algorithm \ref{algorithm:TrajPooler}), which focuses on learning perception features around the predicted trajectory rather than processing the entire image.} given a noisy trajectory $\hat{\tau}^{(m)}$, we first recover its absolute coordinates via cumulative summation. We then sparsely sample 3D points along the trajectory and project them onto the image plane using the known camera intrinsics $\mathbf{K}$ and extrinsics $\mathbf{T}$. Multi-scale image features are extracted at these projected locations from the BEV encoder output. To adaptively aggregate these features, attention weights $\alpha^{(m)}$ are learned conditioned on the initial trajectory feature $F_{\text{traj}}^{(m)}$. The final trajectory-aware representation $F_{\text{traj}}^{(m)}$ is obtained through a weighted summation of the sampled features, followed by an MLP and residual fusion with the original trajectory embedding.

In the second stage, we inject the pooled trajectory token into a transformer-based conditional decoder, as shown in Algorithm \ref{algorithm:DiffusionTrajectoryDecoder}. Specifically, we follow a hierarchical attention stack where the trajectory token $F_{\text{traj}}^{(m)}$ is first used to query agent-level memory via multi-head cross-attention (MHCA), producing agent-conditioned features $F_{\text{agent}}^{(m)}$. These are further refined by attending to map-level memory features, yielding $F_{\text{map}}^{(m)}$. Finally, navigation-level anchors (\textit{e.g.}, waypoints from reference plans or diffusion priors) are fused via another MHCA block. The final output $F_{\text{out}}^{(m)}$ is obtained by feeding the resulting feature through a residual FFN block.
\textcolor{black}{It should be emphasized that the anchor trajectories in DIVER are conceptually different from the reference candidates in VADv2 or Hydra-MDP, which are used for final trajectory selection. In contrast, DIVER employs only six anchors as generative guidance to encourage diverse motion patterns, rather than selecting the final output from them as VADv2 or Hydra-MDP does.}

Together, this conditional feature interaction pipeline ensures that the predicted trajectory at each diffusion step is semantically grounded in both visual context and structural priors, enhancing both diversity and plausibility of multi-mode predictions.

\begin{algorithm}[t]
\caption{TrajPooler}\label{algorithm:TrajPooler}
\KwIn{
    Noisy trajectory $\hat{\tau}^{(m)}\in \mathbb{R}^{T \times 2}$, \\
    Trajectory feature $F_{\text{traj}}^{(m)}\in \mathbb{R}^d$, \\
    Multi-scale image features $\mathcal{F}_{\text{img}}$, \\
    Camera intrinsics $\mathbf{K}$, Extrinsics $\mathbf{T}$
}
\KwOut{Trajectory-aware fused feature $F_{\text{traj}}^{(m)}$}

\textbf{Convert to absolute positions:} $\tau^{(m)}\leftarrow \text{CumulativeSum}(\hat{\tau}^{(m)})$ 

\textbf{Generate 3D keypoints along trajectory:} $P^{(m)}\leftarrow \text{Sample3DPoints}(\tau^{(m)})$ 

\textbf{Project to image planes:} $u^{(m)}\leftarrow \text{ProjectTo2D}(P^{(m)}, \mathbf{K}, \mathbf{T})$ 

\textbf{Sample multi-scale image features:} $F_{2D}^{(m)}\leftarrow \text{SampleImageFeatures}(\mathcal{F}_{\text{img}}, u^{(m)})$ 

\textbf{Learned attention weights:} $\alpha^{(m)}\leftarrow \text{AttentionWeights}(F_{\text{traj}}^{(m)}, \mathbf{K}, \mathbf{T})$

\textbf{Fused trajectory feature:} $F_{\text{traj}}^{(m)}\leftarrow \text{MLP}(\sum \alpha^{(m)}\cdot F_{2D}^{(m)}) + F_{\text{traj}}^{(m)}$
\end{algorithm}

\begin{algorithm}[t]
\caption{Conditional Trajectory Decoder}\label{algorithm:DiffusionTrajectoryDecoder}
\KwIn{

    Trajectory token $F_{\text{traj}}^{(m)}$ \\
    Agent memory tokens $\mathcal{F}_{\text{agent}}$ \\
    Map memory tokens $\mathcal{F}_{\text{map}}$ \\
    Navigation memory $\mathcal{F}^{(m)}{\tau^{\text{a}}}$ \\
    Planning query $Q^{(m)}$
}
\KwOut{Multi-Mode Trajectories:$F_{\text{out}}^{(m)}$}

\textbf{Agent-Level Cross Attention:}

$F_{\text{traj}}^{(m)}= \text{MHCA}(Q^{(m)}+ F_{\text{traj}}^{(m)},\ \mathcal{F}_{\text{agent}}, \mathcal{F}_{\text{agent}})$

\textbf{Map-Level Cross Attention:}
$F_{\text{traj}}^{(m)}= \text{MHCA}(Q^{(m)}+ F_{\text{traj}}^{(m)},\ \mathcal{F}_{\text{map}},\ \mathcal{F}_{\text{map}})$

\textbf{Anchor-Level Cross Attention:}
$F_{\text{traj}}^{(m)}= \text{MHCA}(F_{\text{traj}}^{(m)},\mathcal{F}^{(m)}{\tau^{\text{a}}},\mathcal{F}^{(m)}{\tau^{\text{a}}}$

\textbf{Final LayerNorm + FFN:}
$F_{\text{out}}^{(m)}= \text{FFN}(\text{LayerNorm}(\mathcal{F}^{(m)}{\tau^{\text{a}}}) + \mathcal{F}^{(m)}{\tau^{\text{a}}}) $
\end{algorithm}

\subsubsection{Reference GTs-Guided Multi-Mode Trajectories.}
\noindent\textbf{Planning Head.}
After the final conditional decoding step, the predicted multi-mode trajectories $\hat{\tau}^{(m)}$ are reconstructed from the final trajectory token $F_{\text{out}}^{(m)}$ via a lightweight MLP-based regression head. These predicted trajectories are not generated independently: each is explicitly guided by a GT-based reference trajectory $\tau^{\text{ref}(m)}$, which provides a mode-aligned semantic anchor for generation.

Specifically, during training, each predicted trajectory mode is paired with a GT-derived reference trajectory that encodes distinct motion intents (\textit{e.g.}, turning, yielding, or lane-changing). By injecting the reference into the conditional decoder and using it to shape the context-aware feature fusion, our model learns to align each mode with a feasible, interpretable, and scene-compliant motion pattern.

At inference time, the model samples multiple noise vectors $\epsilon^{(m)}\sim \mathcal{N}(0, I)$ and generates corresponding multi-mode trajectories $\hat{\tau}^{(m)}$ via the denoising diffusion process. The reference trajectories $\tau^{\text{ref}(m)}$ can be drawn from anchor plans, plan queries, or diffusion priors, ensuring that the output space spans diverse yet goal-directed behaviors. \textcolor{black}{Finally, we follow the strategy of SparseDrive  \cite{sun2024sparsedrive} and DiffusionDrive  \cite{liao2024diffusiondrive}, employing a post-processing procedure in which the remaining trajectories are scored based on comfort and scene compliance, and the trajectory with the highest score is selected as the final output.}

\noindent\textbf{Matching Loss for Diverse Trajectories.}
To promote the generation of diverse and plausible trajectories, we introduce a matching loss between predicted multi-mode trajectories and reference GT trajectories. Instead of supervising the model to regress toward a single GT trajectory, we sample reference GT trajectories from the empirical distribution of expert behaviors and match them to the model’s predicted multi-mode trajectories. Specifically:

We denote the predicted trajectory set as ${\hat{\tau}^{(1)}, \dots, \hat{\tau}^{(M)}}$ and the reference GTs set as ${\tau^{\text{ref}(1)}, \dots, \tau^{\text{ref}(M)}}$.
We apply the \textbf{Hungarian algorithm} to compute an optimal one-to-one matching between predicted and reference trajectories by minimizing the pairwise $\ell_2$ distance:

\begin{equation}
    \mathcal{L}_{\mathrm{match}}= \frac{1}{M}\sum_{m=1}^{M}\left\| \hat{\tau}^{(m)}- \tau^{\text{ref}(m)}\right\|_2^2
\end{equation}

This strategy enforces that each predicted mode is assigned to a distinct reference, which directly encourages the diffusion model to explore multiple plausible futures rather than collapsing to a mean or dominant mode.

This reference-guided design not only enhances trajectory diversity by explicitly modeling distinct semantic intents, but also ensures goal feasibility and safety by grounding each mode in plausible motion priors. This is particularly crucial for addressing the mode collapse and over-conservative behaviors that plague prior IL-based E2E-AD methods.

\subsection{Reinforcement Learning for Trajectory Planning}
\subsubsection{Overview of Group Relative Policy Optimization}
\label{sec:grpo}
Group Relative Policy Optimization (GRPO) is a reinforcement learning algorithm tailored for multi-agent or multi-mode scenarios, where agent policies are updated relative to a shared group baseline. Unlike standard PPO  \cite{ppo}, GRPO  \cite{GRPO} formulates the policy gradient as a relative advantage over group-conditioned expectations, thereby encouraging diverse yet cooperative behaviors. The GRPO objective is given by:

\begin{equation}
\mathcal{J}_{\mathrm{GRPO}}(\theta)=\mathbb{E}_{q,\left\{o_{i}\right\}\sim \pi_{\theta_{\textrm{old}}}}\left[\frac{1}{G}\sum_{i=1}^{G}\mathcal{L}_{i}-\beta \mathbb{D}_{K L}\left(\pi_{\theta}\| \pi_{\text {ref }}\right)\right],
\end{equation}
\begin{equation}
\mathcal{L}_{i}=\min \left(w_{i}A_{i}, \operatorname{clip}\left(w_{i}, 1-\epsilon, 1+\epsilon\right) A_{i}\right),
\end{equation}
where, $w_{i}=\frac{\pi_{\theta}\left(o_{i}\mid q\right)}{\pi_{\theta_{\mathrm{old}}}\left(o_{i}\mid q\right)}$, $\epsilon$ and $\beta$ are hyper-parameters, the old policy $\pi_{\theta_{\textrm{old}}}$ and the advantage $A_{i}$ is computed using the normalized reward within the group.

\subsubsection{GRPO for Trajectory Planning}

Despite their strong capability in multi-mode trajectory generation, diffusion models still face two critical limitations in E2E-AD: (i) mode collapse, where generated trajectories converge to similar patterns, and (ii) lack of safety awareness, leading to kinematically feasible yet unsafe motions. These issues largely stem from reliance on single-GT supervision and the absence of explicit guidance for task-specific objectives such as behavioral diversity and safety.

To address this, we treat the diffusion model as a stochastic policy and introduce a reinforcement learning objective based on Group Relative Policy Optimization (GRPO) \cite{GRPO}, which augments the standard imitation learning pipeline with trajectory-level rewards. GRPO enables direct optimization over non-differentiable objectives, facilitates exploration beyond expert demonstrations, and improves policy robustness by aligning generation with safety and diversity constraints. This hybrid optimization paradigm steers the diffusion process toward producing diverse, safe, and goal-directed trajectories that better support the downstream planning requirements of E2E-AD.

\textcolor{black}{
Hydra-MDP  \cite{li2024hydra} promotes multi-mode trajectory generation without relying on imitation learning by directly distilling PDMS-based metrics into the policy network, where such metric distillation can be regarded as a single-step reinforcement learning process. 
    Following the principles of Hydra-MDP  \cite{li2024hydra} and GTRS  \cite{GTRS}, we incorporate a PDMS-guided reinforcement optimization into our framework, implemented on the NAVSIM dataset. 
    Specifically, we reuse the PDMS from Hydra-MDP to compute driving-quality metrics including no at-fault collisions (NC), drivable area compliance (DAC), time-to-collision (TTC), comfort (Comf.) and ego progress (EP), and provide online reinforcement guidance within the GRPO optimization loop. 
    Unlike Hydra-MDP, which distills these metrics into a static policy, our approach dynamically integrates the PDMS-based rewards during training, enabling the diffusion policy to adaptively align with high-level safety and driving-quality objectives. 
    This PDMS-guided reinforcement mechanism complements the trajectory-level rewards of GRPO, thereby enhancing multi-mode diversity, safety awareness, and behavioral consistency.
}

We define a composite reward function $r(\hat{\tau})$ that assesses each trajectory according to \textit{diversity}, \textit{safety}, \textit{trajectory consistency}, and \textit{lane keeping}. 
The detailed formulations of the diversity reward $r_{\text{div}}$, safety reward $r_{\text{safe}}$, trajectory consistency reward $r_{\text{TC}}$, and lane-keeping reward $r_{\text{LK}}$ are presented below.

\noindent \textbf{Diversity Reward $r_{\text{div}}$.}
To encourage multi-mode trajectory generation and mitigate mode collapse, we design a diversity reward $r_{\text{div}}$ that quantifies the pairwise dissimilarity among predicted trajectories. Given a set of $M$ predicted trajectories $\{\hat{\tau}^{(1)}, \hat{\tau}^{(2)}, \dots, \hat{\tau}^{(M)}\}$, we compute:
\begin{align}
    r_{\text{div}}= \frac{2}{M(M-1)}\sum_{i=1}^{M}\sum_{j=i+1}^{M}\| \hat{\tau}^{(i)}- \hat{\tau}^{(j)}\|_2
\end{align}
This reward promotes diverse trajectory generation by maximizing inter-trajectory distances, encouraging the diffusion model to explore a broader set of plausible behaviors rather than collapsing to redundant modes. It is differentiable and seamlessly integrates with GRPO to guide the policy toward both diversity and feasibility.

\noindent \textbf{Safety Reward $r_{\text{safe}}$.} To ensure the feasibility and risk-awareness of generated trajectories, we introduce a safety reward that penalizes predicted paths with low clearance from static or dynamic obstacles. We define a differentiable distance-based cost function using a safety map $D_{\text{safe}}(\mathbf{x})$, which encodes inverse proximity to obstacles. For a trajectory $\hat{\tau}= { \hat{x}\cdot t }\cdot{t=1}^T$, the safety reward is computed as:
\begin{equation}
r_{\text{safe}}(\hat{\tau}) = - \frac{1}{T}\sum_{t=1}^{T}\mathbb{I}\left[ D_{\text{safe}}(\hat{x}\cdot t) < d_{\text{thresh}}\right],
\end{equation}
where $\mathbb{I}[\cdot]$ is an indicator function and $d_{\text{thresh}}$ is a safety margin. This term penalizes any predicted point too close to obstacles, promoting collision-free behaviors. Integrated with the GRPO objective, this safety-aware reward guides the diffusion model to sample physically plausible and safe trajectories in complex, dynamic environments.

\noindent \textcolor{black}{\textbf{Trajectory Consistency Reward $r_{\text{TC}}$.} 
To maintain temporal coherence across consecutive planning cycles, we introduce a trajectory consistency reward that enforces alignment between the current predicted trajectory $\hat{\tau}_t$ and the previous executed (or predicted) trajectory $\hat{\tau}_{t-1}$. Sudden directional shifts between adjacent planning steps may lead to unstable or oscillatory driving behaviors. The reward is formulated as:
\begin{align}
r_{\text{TC}}= - \frac{1}{T}\sum_{k=1}^{T}\left\| \hat{x}_{t,k}- \hat{x}_{t-1,k}\right\|_2,
\end{align}
where $\hat{x}_{t,k}$ denotes the $k$-th waypoint of the trajectory at time $t$. By minimizing inter-frame deviations, this reward encourages smooth transitions and consistent motion patterns over time, reducing unnecessary steering oscillations and improving temporal stability.
}

\noindent \textcolor{black}{\textbf{Lane Keeping Reward $r_{\text{LK}}$.} 
To further constrain the ego vehicle to follow reasonable lane-keeping behaviors, we design a lane keeping reward that penalizes frequent or unnecessary lane changes. Let $\mathcal{L}(\hat{x}_t^{i})$ denote the distance from the $i$-th predicted waypoint $\hat{x}_t^{i}$ to the lane centerline. And $\tau_{LK}$ denotes the distance threshold for determining lane-keeping compliance, which is set to 0.5. We apply the following rule to each waypoint $\hat{x}_t^{i}$ to obtain a binary (0/1) array $arr$.
\begin{align}
arr = [((\mathcal{L}(\hat{x}_t^{i}) - \tau_{LK}) > 0),...] ,\ i= 0,1,...,k,
\end{align}
We iterate through the array $arr$ and compute the maximum $l_{arr}$ length of consecutive ones in the array. Then $r_{\text{LK}}$ can be computed by:
\begin{align}
r_{\text{LK}}= 1 - ((l_{arr}- \delta_{LK}) > 0),
\end{align}
where $\delta_{LK}$ represents the threshold and is set to 4.
}

\noindent \textbf{Planning Loss for Diverse and Consistent Trajectories.}
\textcolor{black}{The total reward for a predicted trajectory $\hat{\tau}^{(m)}$ is defined as:
\begin{align}
r(\hat{\tau}^{(m)}) =\;& 
\lambda_{\text{div}}\cdot r_{\text{div}}(\hat{\tau}^{(m)}) 
+ \lambda_{\text{safe}}\cdot r_{\text{safe}}(\hat{\tau}^{(m)}) \nonumber \\
& + \lambda_{\text{TC}}\cdot r_{\text{TC}}(\hat{\tau}^{(m)}) 
+ \lambda_{\text{LK}}\cdot r_{\text{LK}}(\hat{\tau}^{(m)}),
\end{align}
where $r_{\text{div}}$, $r_{\text{safe}}$, $r_{\text{TC}}$, and $r_{\text{LK}}$ denote the diversity, safety, trajectory consistency, and lane-keeping rewards, respectively.
It is worth noting that \textbf{DIVER} is primarily designed to mitigate trajectory mode collapse and enhance trajectory diversity. Accordingly, DIVER incorporates the reward terms $r_{\text{TC}}$ and $r_{\text{LK}}$. In the ablation study, we further compare the effects of $r_{\text{TC}}$, $r_{\text{LK}}$, and PDMS-guided strategies. Overall, our DIVER framework provides a scalable and extensible reward design, which can be flexibly adapted or revised to align with different objectives and human values.
}

The total reward is integrated into a reinforcement learning loss under the GRPO framework:
\begin{align}
\mathcal{L}_{\text{RL}}= -\mathbb{E}_{\hat{\tau}^{(m)}\sim \pi_{\theta}}
\left[ \text{GRPO}\big(\hat{\tau}^{(m)}, r(\hat{\tau}^{(m)})\big) \right].
\end{align}

The complete training objective combines imitation learning and reinforcement learning:
\begin{equation}
\mathcal{L}_{\text{total}}= 
\lambda_{\text{match}}\mathcal{L}_{\text{match}}
+ \lambda_{\text{RL}}\mathcal{L}_{\text{RL}},
\end{equation}
where $\lambda_{\text{match}}$ and $\lambda_{\text{RL}}$ are weighting coefficients balancing the contributions of trajectory matching and policy optimization.

Our reinforcement-augmented training framework addresses key limitations of diffusion-based planners in E2E-AD. 
Through reward-guided GRPO optimization, we impose safety, diversity, and trajectory consistency constraints that are not captured by conventional MSE-based training. 
This hybrid supervision strategy bridges the gap between behavior cloning and policy-level optimization, leading to more stable, safe, and deployable planning performance.

\section{Experiments}\label{sec:exps}
\subsection{Dataset}

\noindent \textbf{Bench2Drive  (Close-Loop).}
We conduct training and evaluation of DIVER on the \textbf{Bench2Drive} \cite{jia2024bench2drive}, a closed-loop evaluation protocol based on the CARLA Leaderboard 2.0 \cite{CARLA} for E2E-AD. It provides a base training set of 1000 clips, with 950 used for training and 50 for open-loop validation. Each clip captures approximately 150 meters of continuous driving in a specific traffic scenario. For closed-loop evaluation, we use the official 220 routes, covering 44 interactive scenarios with 5 routes each.

\noindent \textbf{NAVSIM  (Close-Loop).}
We conduct training and evaluation of DIVER on the \textbf{NAVSIM} \cite{dauner2024navsim} dataset. NAVSIM is a real-world, planning-oriented dataset that builds upon OpenScene, a compact redistribution of nuPlan \cite{caesar2021nuplan}, the largest publicly available annotated driving dataset. It leverages eight cameras to achieve a full 360° field of view, along with a merged LiDAR point cloud derived from five sensors. Annotations are provided at 2 Hz and include both HD maps and object bounding boxes. The dataset is specifically designed to emphasize challenging driving scenarios involving dynamic changes in driving intentions, while deliberately excluding trivial cases such as stationary scenes or constant-speed cruising.

\noindent \textbf{NuScenes (Open-Loop).}
We conduct extensive open-loop experiments on the \textbf{nuScenes} dataset \cite{nuscenes}, which consists of 1000 driving scenes (700 for training, 150 for validation and 150 for test). Each scene lasts 20 seconds and includes around 40 key-frames annotated at 2 Hz. Each sample contains six images from surround-view cameras (covering 360° FOV), and point clouds from both LiDAR and radar sensors. 

\noindent \textbf{Turning-nuScenes (Open-Loop).}
We conduct extensive open-loop experiments on the \textbf{Turning-nuScenes} dataset \cite{momad}, a challenging subset of NuScenes proposed by MomAD \cite{momad} to evaluate trajectory consistency in non-trivial maneuvers. While most planning tasks in the original nuScenes dataset primarily involve go-straight commands, Turning-nuScenes specifically focuses on turning scenarios to assess the temporal coherence of predicted trajectories. To construct this subset, samples are selected based on a displacement threshold of 25 meters between the predicted positions at 0.5s and 3.0s in the GT ego trajectory. The resulting validation set comprises 680 samples across 17 scenes, accounting for approximately one-tenth of the full nuScenes validation set.

\noindent \textbf{Adv-nuSc  (Open-Loop).}
To evaluate adversarial robustness, we conduct extensive open-loop experiments on the \textbf{Adv-nuSc} \cite{xu2025challenger} dataset. It contains 156 scenes (6,115 samples) and is specifically crafted to challenge the ego vehicle by introducing adversarial traffic participants.It is built upon the validation split of the nuScenes dataset  \cite{nuscenes}, which contains 150 scenes, each with 20 seconds of driving data. For each scene, we randomly select up to 10 background vehicles (if there are that many) that come close to the ego vehicle at any point in time and designate them as candidate adversarial agents. Challenger is then used to generate adversarial trajectories for these vehicles, creating diverse and challenging driving scenarios.

\noindent \textbf{NuScenes-C  (Open-Loop).}
\textbf{NuScenes-C} \cite{zhujun_benchmarking} is a corrupted benchmark derived from the nuScenes validation set, introducing various types of noise to assess the robustness of planning models. It includes 27 corruption types applied at 5 severity levels. To evaluate robustness under adverse weather conditions, we select three representative weather corruptions — Rain, Snow, and Fog — as our test scenarios.

\subsection{Evaluation Metrics}
\noindent\textbf{Close-Loop (Bench2Drive).} The Bench2Drive \cite{jia2024bench2drive} includes five metrics for closed-loop evaluation: Driving Score (DS), Success Rate (SR), Efficiency, Comfortness, and Multi-Ability. The Success Rate quantifies the proportion of routes successfully completed within the allotted time. The Driving Score follows CARLA [11], incorporating both route completion status and violation penalties, where infractions reduce the score via discount factors. Efficiency and Comfortness are used to measure the speed performance and comfort of the autonomous driving system during the driving process, respectively. Multi-Ability measures 5 advanced skills, including `Merging, Overtaking, Emergency Brake, Give Way, and Traffic Sign', independently for urban driving. 

\noindent\textbf{Close-Loop (NAVSIM v1).}
NAVSIM v1\cite{dauner2024navsim} metrics include No at-fault Collision (NC), Drivable Area Compliance (DAC), Time-to-Collision (TTC), Comfort (C.), and Ego Progress (EP). NAVSIM uses the Predictive Driver Model Score (PDMS) to evaluate model performance.

\noindent\textbf{Close-Loop (NAVSIM v2).}
NAVSIM v2~\cite{navsimv2} evaluates planners using the Extended Predictive Driver Model Score (EPDMS), which aggregates penalties and weighted subscores including no at-fault collision (NC), drivable area compliance (DAC), driving direction compliance (DDC), traffic light compliance (TLC), ego progress (EP), time to collision (TTC), lane keeping (LK), history comfort (HC), and extended comfort (EC). It also introduces a \textit{Navhard} split with a two-stage protocol: (i) the planner is scored on real observations, and (ii) it is re-evaluated under perturbed future observations synthesized via 3D Gaussian Splatting around the stage-1 endpoint. The final EPDMS is obtained by Gaussian-weighted aggregation of the two-stage scores.

\noindent\textbf{Open-Loop.} For the nuScenes dataset \cite{nuscenes}, Adv-nuSc \cite{xu2025challenger},  Turning-nuScenes \cite{momad}, and nuScenes-C \cite{zhujun_benchmarking}, we adopt the commonly used collision rate to assess planning performance, as in SparseDrive \cite{sun2024sparsedrive}.
We argue that the \textcolor{red}{\textbf{L2 distance}} is \textcolor{red}{\underline {not}} a suitable evaluation metric for multi-mode E2E-AD methods, as it only measures proximity to a single GT trajectory and fails to capture the benefits of multi-mode prediction.
To address this, we introduce a new \textbf{\textit{Diversity Metric}} (denoted as \textbf{Div.}) to better evaluate the variety and richness of predicted trajectories. This metric is designed to be scale-invariant, modality-invariant, and bounded within \([0, 1]\), making it suitable for consistent comparison across models and scenarios.

Let \( M \) be the number of predicted modes (e.g., \( M = 6 \)), and \( T \) the number of waypoints per trajectory (e.g., \( T = 6 \)). Each trajectory consists of 2D coordinates, with \( \mathbf{p}_t^{(i)}= (x_t^{(i)}, y_t^{(i)}) \) denoting the \( t \)-th waypoint of the \( i \)-th trajectory.

To characterize the temporal variation in trajectory diversity, we compute a time-conditioned diversity score at each future timestamp. For example, in the Bench2Drive and nuScenes datasets, we consider  $t \in \{0.5, 1.0, 1.5, 2.0, 2.5, 3.0\}$ seconds.
We first compute the unnormalized pairwise diversity as:
\begin{align}
    D_{\text{raw}}^{(t)}= \frac{2}{M(M-1)}\sum_{i=1}^{M-1}\sum_{j=i+1}^{M}\left\| \mathbf{p}_{t}^{(i)}- \mathbf{p}_{t}^{(j)}\right\|_2.
\end{align}
To ensure comparability across scenes with varying trajectory scales, we normalize it using the mean trajectory magnitude:

\begin{align}
 \textit{Div.}^{(t)}= \min\left(1, \frac{D_{\text{raw}}^{(t)}}{\epsilon + \frac{1}{M}\sum_{m=1}^{M}\left\| \mathbf{p}_{t}^{(m)}\right\|_2}\right),
\end{align}
where \( \epsilon \) is a small constant (\textit{e.g.}, \( 10^{-6}\)) to avoid division by zero.

Overall, a higher \textbf{Diversity} value indicates greater dispersion among the predicted trajectories, reflecting a richer coverage of intent modes such as lane changes, braking, and yielding. When Diversity approaches 1, the predicted modes are highly diverse and well-separated in the trajectory space. Conversely, a value near 0 suggests that the predicted trajectories are highly similar, indicating a lack of behavioral diversity. Notably, the Diversity Metric is only applicable to multi-mode trajectory E2E-AD methods.
\subsection{Implementation Details}
\label{sec:Implementation_Details}

\begin{table*}[t]
\scriptsize
\centering
\addtolength{\tabcolsep}{0.1pt}
\caption{$\operatorname{ \textbf{Open-Loop}}$, $\operatorname{\textbf{Closed-Loop}}$ results and $\operatorname{Multi-Ability}$ results on \textbf{Bench2Drive} (V0.0.3) under base training set.
`mmt' refers multi-mode trajectory variant of $\operatorname{VAD}$ and $^\dagger$ denotes the re-implementation. * denotes expert feature distillation. `DS' denotes Driving Score. `SR' denotes Success Rate. `Effi' denotes Efficiency. `Comf' denotes Comfortness. `Merg.' denotes Merging.  `Overta.' denotes Overtaking. `Emerge.' denotes Emergency Brake. It is worth noting that the \textbf{\textit{Diversity Metric}}($\operatorname{ \textit{Div.}^{(t)}}\uparrow$) is applicable only to multi-mode E2E-AD methods. 
}

  \renewcommand\arraystretch{0.7}
  \tabcolsep=0.1mm 
  \resizebox{\linewidth}{!}{
  \begin{tabular}{lcc c cc cccc cccccc}
    \toprule
   \multirow{2}{*}{$\operatorname{Method}$}&\multirow{2}{*}{$\operatorname{Traj.}$}&\multirow{2}{*}{$\operatorname{Scheme}$}& \multirow{2}{*}{$\operatorname{Venue}$}&  \multicolumn{2}{c}{$\operatorname{Open-loop\ Metric}$}    & \multicolumn{4}{c}{$\operatorname{Closed-loop\ Metric}$}& \multicolumn{6}{c}{$\operatorname{Multi-Ability (\%) \uparrow}$}  \\
   \cmidrule(lr){5-6}\cmidrule(lr){7-10}\cmidrule(lr){11-16}
   &&&&$\operatorname{Avg. L2}\downarrow$&$\operatorname{Avg. \textit{Div.}^{(t)}}\uparrow$&$\operatorname{DS}\uparrow$ & $\operatorname{SR\ (\%)}\uparrow$ & $\operatorname{Effi}\uparrow$  & $\operatorname{Comf}\uparrow$ &
   $\operatorname{Merg.}$& $\operatorname{Overta.}$& $\operatorname{Emerge.}$& $\operatorname{Give\:Way}$ &$\operatorname{Traffic\:Sign}$& $\operatorname{Mean}$
   \\
   
\midrule
ThinkTwice*  \cite{thiktwice}&ST& IL&CVPR 2023&0.95& -& 62.44& 31.23 &69.33& 16.22& 27.38&18.42& 35.82& \textbf{50.00}& 54.23& 37.17\\
DriveAdapter* \cite{jia2023driveadapter}&ST& IL&ICCV 2023 & 1.01&-& 64.22& 33.08 &70.22& 16.01& 28.82&26.38& 48.76& \textbf{50.00}& 56.43& 42.08\\
DriveTrans* \cite{jia2025drivetransformer}&MT& IL&ICLR 2025&\textbf{0.62}&- & 63.46& 35.01 &\textbf{100.64}& 20.78& 17.57& 35.00& 48.36& 40.00& 52.10& 38.60\\
WoTE* \cite{wote}& MT& IL &ICCV 2025&-&-  &61.71& 31.36 &-& -& -& -& -& -& -& -\\
\textcolor{black}{DiffAD* \cite{wang2025diffad}}& MT & IL &Arxiv 2025&1.55&-  &67.92& \textbf{38.64}&- &- &30.00& \textbf{35.55}& \textbf{46.66}& 40.00& 46.32& 38.79\\
\midrule
${\operatorname{ThinkTwice}_{\operatorname{mmt}}}^{*\dagger }$ \cite{thiktwice}&MT& IL&CVPR 2023&0.93& 0.19& 63.34& 33.23 &71.56& 18.32& 31.31&21.23& 38.33& \textbf{50.00}& 57.45& 39.66\\
\cellcolor{gray!15}${\operatorname{DIVER}}$ (Ours)&\cellcolor{gray!15}MT& \cellcolor{gray!15}IL\&RL& \cellcolor{gray!15}-& \cellcolor{gray!15}1.11& \cellcolor{gray!15}\textbf{0.38}& \cellcolor{gray!15}\textbf{68.90}& \cellcolor{gray!15}36.75 & \cellcolor{gray!15}72.34& \cellcolor{gray!15}\textbf{22.34}& \cellcolor{gray!15}\textbf{35.08}& \cellcolor{gray!15}25.09& \cellcolor{gray!15}41.09& \cellcolor{gray!15}\textbf{50.00}& \cellcolor{gray!15}\textbf{59.21}& \cellcolor{gray!15}\textbf{42.09}\\
\midrule
\midrule
UniAD-Base \cite{uniad}&ST& IL&CVPR 2023&\textbf{0.73}&- & 45.81& 16.36& 129.21& 43.58& 14.10& 17.78& 21.67& 10.00& 14.21& 15.55\\
$\operatorname{VAD}$ \cite{jiang2023vad}&ST&IL &ICCV 2023& 0.91&-& 42.35& 15.00& 157.94& 46.01&8.11& 24.44& 18.64& \textbf{20.00}& 19.15& 18.07 \\
GenAD \cite{zheng2024genad}&ST& IL&ECCV 2024&-&- &44.81& 15.90& -& -& -& -& -& -& -& -\\
$\operatorname{MomAD(VAD)}$ \cite{momad}&MT&IL&CVPR 2025&0.87 & 0.18& 45.35& 17.44& 162.09& 49.34&9.99& 26.31& 20.07& \textbf{20.00}& 20.23&19.32 \\
$\operatorname{MomAD(SD)}$ \cite{momad}&MT&IL&CVPR 2025&0.82 & 0.20 &47.91& 18.11& 174.91 &51.20&13.21& 21.02& 18.01& \textbf{20.00}& 21.07&18.66  \\
\midrule
${\operatorname{VAD}_{\operatorname{mmt}}}^{\dagger }$ \cite{jiang2023vad}&MT&IL &ICCV 2023&0.89& 0.20& 42.87& 15.91& 158.12& 47.22&9.43& 25.31& 19.91& \textbf{20.00}& 20.09&18.95   \\
\cellcolor{gray!15}$\operatorname{DIVER\ (Ours)}$&\cellcolor{gray!15}MT&\cellcolor{gray!15}IL\&RL&\cellcolor{gray!15}-&\cellcolor{gray!15}1.13 & \cellcolor{gray!15}0.32& \cellcolor{gray!15}47.95& \cellcolor{gray!15}19.47 & \cellcolor{gray!15}164.66 & \cellcolor{gray!15}51.28 & \cellcolor{gray!15}13.83 & \cellcolor{gray!15}\textbf{29.09} & \cellcolor{gray!15}\textbf{25.51}& \cellcolor{gray!15}\textbf{20.00} & \cellcolor{gray!15}\textbf{24.93} & \cellcolor{gray!15}\textbf{22.67} \\
\midrule
${\operatorname{SparseDrive}}^{\dagger}$ \cite{sun2024sparsedrive}&MT&IL&ICRA 2025&0.87&  0.21& 44.54& 16.71& 170.21& 48.63&12.18& 23.19& 17.91& \textbf{20.00}& 20.98& 18.85 \\

\cellcolor{gray!15}$\operatorname{DIVER\ (Ours)}$  &\cellcolor{gray!15}MT&\cellcolor{gray!15}IL\&RL&\cellcolor{gray!15}-&\cellcolor{gray!15}1.05&\cellcolor{gray!15}\textbf{0.35}&\cellcolor{gray!15}\textbf{49.21}&\cellcolor{gray!15}\textbf{21.56}&\cellcolor{gray!15}\textbf{177.00}&\cellcolor{gray!15}\textbf{54.72}&\cellcolor{gray!15}\textbf{15.98} &\cellcolor{gray!15}28.22 &\cellcolor{gray!15}23.71 &\cellcolor{gray!15}\textbf{20.00}&\cellcolor{gray!15}24.38 &\cellcolor{gray!15}22.46\\
\bottomrule
\end{tabular}}
\label{tab_b2d}
\end{table*}

\begin{table}[t]
\centering
  \caption{Comparison on \textbf{NAVSIM v1} navtest split with $\operatorname{\textbf{Closed-Loop}}$ metrics. `mmt' refers multi-mode trajectory variant of $\operatorname{TransFuser}$ and $^*$ denotes the re-implementation. }
\renewcommand\arraystretch{0.7}
  \tabcolsep=0.8mm 
  \resizebox{\linewidth}{!}{
\begin{tabular}{l c cccccc}
\toprule
\multicolumn{1}{l}{Method}& \multicolumn{1}{c}{Input} & NC$\uparrow$                & \multicolumn{1}{c}{DAC}$\uparrow$ & TTC$\uparrow$                 & Comf.$\uparrow$                & EP$\uparrow$                  & PDMS$\uparrow$ \\
\midrule
$\operatorname{UniAD}$  \cite{uniad}& C & 97.8 & 91.9 & 92.9 & \textbf{100} & 78.8 &83.4 \\
$\operatorname{LAW}$  \cite{li2024enhancing}& C & 97.8 & 91.9 & 92.9 & \textbf{100} & 78.8 &83.4 \\
$\operatorname{LTF}$ \cite{TransFuser}& C & 97.4 & 92.8 & 92.4 & \textbf{100} & 79.0 & 83.8 \\
$\operatorname{PARA-Drive}$ \cite{paradrive}& C & 97.9 & 92.4 & 93.0 & 99.8 & 79.3 & 84.0 \\
\textcolor{black}{$\operatorname{DiffRefiner}$ \cite{yin2025diffrefiner}}& C& 98.4& 97.4& 95.3 &\textbf{100}& 83.4 & 89.4 \\
$\operatorname{VADv2}$ \cite{chen2024vadv2}& C\&L& 97.2 & 89.1 & 91.6 & \textbf{100} & 76.0 & 80.9 \\
\textcolor{black}{$\operatorname{Hydra-MDP}$ \cite{li2024hydra}}& C\&L& 98.3 & 96.0 & 94.6 & \textbf{100} & 78.7 & 86.5 \\
\textcolor{black}{$\operatorname{TrajHF}$ \cite{TrajHF}}& C\&L& 96.3& 96.0 & 91.5 & \textbf{100} & 83.1 & 86.4 \\
\textcolor{black}{$\operatorname{DriveSuprim}$ \cite{yao2025drivesuprim}}& C\&L& 97.8& 97.3 &93.6  &\textbf{100}  & \textbf{86.7}& 89.9 \\
\textcolor{black}{$\operatorname{DiffusionDrive}$ \cite{liao2024diffusiondrive}}& C\&L& 98.2 & 96.2 & 94.7 & \textbf{100} & 82.2 & 88.1 \\
\textcolor{black}{$\operatorname{BridgeDrive}$ \cite{BridgeDrive}}& C\&L& 98.2 &96.1 &94.5& \textbf{100} &82.3& 88.0 \\
\textcolor{black}{$\operatorname{GoalFlow}$ \cite{xing2025goalflow}}& C\&L& 98.4 &\textbf{98.3} & 94.6 & \textbf{100} & 85.0 & \textbf{90.3} \\
\textcolor{black}{$\operatorname{DiffE2E}$ \cite{zhaodiffe2e}}& C\&L& \textbf{99.2}& 96.8 &\textbf{96.7} &\textbf{100} &83.6& 89.8 \\
$\operatorname{TransFuser}$ \cite{TransFuser}& C\&L& 97.7 & 92.8 & 92.8 & \textbf{100} & 79.2 & 84.0 \\ 
${\operatorname{TransFuser}_{\operatorname{mmt}}}^{*}$ \cite{TransFuser}& C\&L&96.2& 95.4& 90.7& \textbf{100}& 80.7& 85.1 \\ 
\cellcolor{gray!15}$\operatorname{DIVER\ (Ours)}$ & \cellcolor{gray!15}C\&L &  \cellcolor{gray!15}98.5 & \cellcolor{gray!15}96.5 & \cellcolor{gray!15}94.9 & \cellcolor{gray!15}\textbf{100} & \cellcolor{gray!15}82.6 & \cellcolor{gray!15}88.3 \\ 
\bottomrule
\end{tabular}}
\label{tab_navsimv1}
\end{table}

\begin{table}[t]
\centering
  \caption{Comparison on \textbf{NAVSIM v2} navtest split with extended  $\operatorname{\textbf{Closed-Loop}}$ metrics. `mmt' refers multi-mode trajectory variant of $\operatorname{TransFuser}$ and $^*$ denotes the re-implementation. }
\renewcommand\arraystretch{0.7}
  \tabcolsep=0.3mm 
  \resizebox{\linewidth}{!}{
\begin{tabular}{l ccccc ccccc}
\toprule
\multicolumn{1}{l}{Method}&  NC$\uparrow$& DAC$\uparrow$& DDC$\uparrow$& TL$\uparrow$& EP$\uparrow$& TTC$\uparrow$& LK$\uparrow$& HC$\uparrow$& EC$\uparrow$& EPDMS$\uparrow$ \\
\midrule
${\operatorname{TransFuser}}$\cite{TransFuser} & 96.9 &89.9& 97.8& 99.7& 87.1& 95.4& 92.7 &98.3 &87.2& 76.7\\
${\operatorname{Hydra-MDP++}}$ \cite{li2025hydraplusplus}& 97.2& \textbf{97.5}& \textbf{99.4} &99.6& 83.1& 96.5& 94.4& 98.2& 70.9& 81.4\\
\textcolor{black}{${\operatorname{DiffusionDriveV2}}$ \cite{zou2025diffusiondrivev2}}& 97.7& 96.6& 99.2& \textbf{99.8}& \textbf{88.9}& 97.2& 96.0& 97.8& \textbf{91.0}& \textbf{85.5}\\
${\operatorname{DriveSuprim}}$\cite{yao2025drivesuprim}& 97.5& 96.5& \textbf{99.4} &99.6 &88.4& 96.6 &95.5& 98.3 &77.0& 83.1\\
${\operatorname{ARTEMIS}}$\cite{feng2025artemis}& \textbf{98.3}& 95.1& 98.6& \textbf{99.8}& 81.5& \textbf{97.4}& \textbf{96.5}& 98.3& -& 83.1\\
${\operatorname{TransFuser}_{\operatorname{mmt}}}^{*}$ \cite{TransFuser}&  97.0 &90.1& 97.6& 99.5& 87.3& 95.5& 92.4 &98.2 &87.3& 76.9\\
\cellcolor{gray!15}$\operatorname{DIVER\ (Ours)}$& \cellcolor{gray!15}97.5 &\cellcolor{gray!15}95.0& \cellcolor{gray!15}98.3& \cellcolor{gray!15}\textbf{99.8}& \cellcolor{gray!15}87.8& \cellcolor{gray!15}95.9& \cellcolor{gray!15}93.0 &\cellcolor{gray!15}\textbf{98.5} &\cellcolor{gray!15}87.7& \cellcolor{gray!15}82.2\\
\bottomrule
\end{tabular}}
\label{tab_navsimv2}
\end{table}

\begin{table}[t]
\centering
  \caption{Comparison on \textbf{NAVSIM v2} navhard split with extended  $\operatorname{\textbf{Closed-Loop}}$ metrics.  }
\renewcommand\arraystretch{0.7}
  \tabcolsep=0.3mm 
  \resizebox{\linewidth}{!}{
\begin{tabular}{l c ccccc ccccc}
\toprule
\multicolumn{1}{l}{Method}&Stage&  NC$\uparrow$& DAC$\uparrow$& DDC$\uparrow$& TL$\uparrow$& EP$\uparrow$& TTC$\uparrow$& LK$\uparrow$& HC$\uparrow$& EC$\uparrow$& EPDMS$\uparrow$ \\
\midrule
\multirow{2}{*}{${\operatorname{LTF}}$\cite{TransFuser}} &Stage 1& 96.2& 79.5& 99.1& 99.5& 84.1& 95.1& 94.2& 97.5& 79.1&\multirow{2}{*}{23.1}\\
 &Stage 2& 77.7& 70.2& 84.2& 98.0& 85.1& 75.6 &45.4& 95.7& 75.9\\
 \multirow{2}{*}{${\operatorname{DiffusionDriv}}$\cite{liao2024diffusiondrive}} &Stage 1& 96.0& 79.7& 97.4 &99.5 &81.3 &93.1& 90.8 &96.8 &73.8&\multirow{2}{*}{24.2}\\
 &Stage 2& 82.1& 72.2& 88.5& 98.7& 85.1& 78.8& 49.2& 89.3& 71.2\\
\multirow{2}{*}{${\operatorname{GuideFlow}}$\cite{liu2025guideflow}} &Stage 1& 96.6& 80.5& 96.3& 99.3& 82.3& 94.9 &91.5& 97.7& 67.8&\multirow{2}{*}{27.1}\\
 &Stage 2& 87.3& 76.7& 88.8& 99.2 &84.3 &85.1 &49.7& 93.1 &44.5\\
\multirow{2}{*}{${\operatorname{MindDrive}}$\cite{liu2025guideflow}} &Stage 1& 96.1& 86.0& 98.8& 99.3 &83.3& 95.6& 94.4 &97.6 &74.7&\multirow{2}{*}{30.5}\\
 &Stage 2& 82.6 &79.1 &86.4 &98.0& 85.3& 79.4 &49.2& 96.5& 71.0\\
\multirow{2}{*}{$\operatorname{DIVER\ (Ours)}$}&Stage 1& 96.4& 80.2& 96.0& 98.8& 82.1& 94.2& 90.6 &96.5& 65.2& \multirow{2}{*}{26.8}\\
&Stage 2& 86.4& 74.9& 87.9& 98.7& 84.1& 84.5& 48.8& 92.6& 43.4\\
\bottomrule
\end{tabular}}
\label{tab_navsimv2_navhard}
\end{table}

\begin{table}[t]
\centering
  \caption{Planning results on the $\operatorname{ \textbf{Open-Loop}}$ \textbf{NuScenes} \cite{nuscenes} validation dataset. It is worth noting that the \textbf{\textit{Diversity Metric}}($\operatorname{\textit{Div.}^{(t)}}\uparrow$) is applicable only to multi-mode E2E-AD methods like MomAD  \cite{momad}, DiffusionDrive  \cite{liao2024diffusiondrive}, and SparseDrive  \cite{sun2024sparsedrive}. 
  }
\renewcommand\arraystretch{0.7}
  \tabcolsep=1.3mm 
  \resizebox{\linewidth}{!}{
  \begin{tabular}{l cccc cccc}
\toprule
\multirow{2}{*}{$\operatorname{Method}$}& \multicolumn{4}{c}{$\operatorname{\textit{Div.}^{(t)}}\uparrow$}& \multicolumn{4}{c}{$\operatorname{Col.\ Rate\ (\%)}\downarrow$}\\
\cmidrule(lr){2-5}\cmidrule(lr){6-9}
& 1s & 2s & 3s & $\operatorname{Avg.}$ & 1s & 2s & 3s & $\operatorname{Avg.}$ \\
\midrule
$\operatorname{UniAD}$ \cite{uniad}&- & - & - & -& 0.62 & 0.58 & 0.63 & \cellcolor{gray!15}0.61  \\
$\operatorname{VAD}$ \cite{jiang2023vad}&- & - & - & -& 0.03 & 0.19 & 0.43 & \cellcolor{gray!15}0.21  \\
\textcolor{black}{$\operatorname{VDT-Auto}$ \cite{guo2025vdt}}&- & - & - & -& 0.05& 0.18& 0.40 & \cellcolor{gray!15}0.21  \\
\textcolor{black}{$\operatorname{Epona}$ \cite{zhang2025epona}}&- & - & - & -& \textbf{0.01}& 0.22& 0.85 &\cellcolor{gray!15}0.36  \\

$\operatorname{MomAD}$ \cite{momad}&   0.05 & 0.10 & 0.19 & \cellcolor{gray!15}0.11 & \textbf{0.01}& \textbf{0.05}& 0.22 & \cellcolor{gray!15}0.09 \\
\textcolor{black}{$\operatorname{DiffusionDrive}$ \cite{liao2024diffusiondrive}}& 0.07& 0.14& 0.24& \cellcolor{gray!15}0.15& 0.03& \textbf{0.05}& 0.16& \cellcolor{gray!15}0.08\\

$\operatorname{SparseDrive}$ \cite{sun2024sparsedrive}&   0.05 & 0.11 & 0.23 & \cellcolor{gray!15}0.13 & \textbf{0.01}& \textbf{0.05}& 0.18 & \cellcolor{gray!15}0.08 \\
\rowcolor{gray!15}$\operatorname{DIVER\ (Ours)}$ &   \textbf{0.10}&  \textbf{0.19}&  \textbf{0.34}& \cellcolor{gray!15}\textbf{0.21}& \textbf{0.01}& \textbf{0.05}& \textbf{0.15}& \cellcolor{gray!15}\textbf{0.07}\\
\bottomrule
\end{tabular}}
\label{tab_nuscenes_planning}
\end{table}

\begin{table*}[t]
\scriptsize
\centering
\caption{Robustness study of planning performance on \textbf{nuScenes} (validation, 6s-horizon; trained for 10 epochs), \textbf{Turning-nuScenes} \cite{momad} (validation), \textbf{Adv-nuSc} (adversarial scenarios), and \textbf{nuScenes-C}. $^*$ denotes the re-implementation.}
\label{tab:robustness_study}
\renewcommand\arraystretch{0.7}
\setlength{\tabcolsep}{0.85mm}
\resizebox{\linewidth}{!}{
\begin{tabular}{l ccc ccc ccc ccc ccc ccc ccc ccc ccc}
\toprule
\multirow{3}{*}{$\operatorname{Method}$} &
\multicolumn{6}{c}{\textbf{nuScenes (6s-horizon)}} &
\multicolumn{6}{c}{\textbf{Turning-nuScenes}} &
\multicolumn{3}{c}{\textbf{Adv-nuSc}} &
\multicolumn{9}{c}{\textbf{nuScenes-C (Col. (\%)$\downarrow$)}} \\
\cmidrule(lr){2-7}\cmidrule(lr){8-13}\cmidrule(lr){14-16}\cmidrule(lr){17-25}
& \multicolumn{3}{c}{$\operatorname{\textit{Div.}^{(t)}}\uparrow$} &
\multicolumn{3}{c}{$\operatorname{Col.\  (\%)}\downarrow$} &
\multicolumn{3}{c}{$\operatorname{\textit{Div.}^{(t)}}\uparrow$} &
\multicolumn{3}{c}{$\operatorname{Col.\  (\%)}\downarrow$} &
\multicolumn{3}{c}{$\operatorname{Col.\  (\%)}\downarrow$} &
\multicolumn{3}{c}{$\operatorname{Snow}$} &
\multicolumn{3}{c}{$\operatorname{Rain}$} &
\multicolumn{3}{c}{$\operatorname{Fog}$} \\
\cmidrule(lr){2-4}\cmidrule(lr){5-7}
\cmidrule(lr){8-10}\cmidrule(lr){11-13}
\cmidrule(lr){14-16}
\cmidrule(lr){17-19}\cmidrule(lr){20-22}\cmidrule(lr){23-25}
& 4s & 5s & 6s & 4s & 5s & 6s &
1s & 2s & 3s & 1s & 2s & 3s &
1s & 2s & 3s &
1s & 2s & 3s &
1s & 2s & 3s &
1s & 2s & 3s \\
\midrule
$\operatorname{UniAD}$ \cite{uniad} &
-- & -- & -- & -- & -- & -- &
-- & -- & -- & -- & -- & -- &
0.800 & 4.100 & 6.960 &
-- & -- & -- &
-- & -- & -- &
-- & -- & -- \\

$\operatorname{VAD}$ \cite{jiang2023vad} &
-- & -- & -- & -- & -- & -- &
-- & -- & -- & -- & -- & -- &
4.460 & 7.590 & 9.080 &
-- & -- & -- &
-- & -- & -- &
-- & -- & -- \\

$\operatorname{SparseDrive}$ \cite{sun2024sparsedrive} &
0.35 & 0.46 & 0.59 & 0.87 & 1.54 & 2.33 &
0.09 & 0.18 & 0.36 & 0.04 & 0.17 & 0.98 &
\textbf{0.029} & 0.618 & 2.430 &
0.13 & 0.27 & 0.50 &
0.11 & 0.27 & 0.55 &
0.14 & 0.36 & 0.58 \\

$\operatorname{DiffusionDrive}^{*}$ \cite{liao2024diffusiondrive} &
0.36 & 0.47 & 0.62 & 0.84 & 1.46 & 2.19 &
0.11 & 0.21 & 0.37 & \textbf{0.03} & 0.14 & 0.85 &
0.068 & 1.299 & 3.646 &
0.09 & 0.24 & 0.39 &
0.07 & 0.18 & 0.35 &
0.06 & 0.18 & 0.30 \\

$\operatorname{MomAD}$ \cite{momad} &
0.33 & 0.43 & 0.52 & 0.83 & 1.43 & 2.13 &
0.09 & 0.17 & 0.34 & \textbf{0.03} & 0.13 & 0.79 &
-- & -- & -- &
0.08 & 0.16 & 0.30 &
0.06 & 0.17 & 0.31 &
0.06 & 0.19 & 0.32 \\

\cellcolor{gray!15}$\operatorname{DIVER\ (Ours)}$ &
\cellcolor{gray!15}\textbf{0.50} & \cellcolor{gray!15}\textbf{0.61} & \cellcolor{gray!15}\textbf{0.75} &
\cellcolor{gray!15}\textbf{0.76} & \cellcolor{gray!15}\textbf{1.32} & \cellcolor{gray!15}\textbf{1.91} &
\cellcolor{gray!15}\textbf{0.17} & \cellcolor{gray!15}\textbf{0.29} & \cellcolor{gray!15}\textbf{0.47} &
\cellcolor{gray!15}\textbf{0.03} & \cellcolor{gray!15}\textbf{0.11} & \cellcolor{gray!15}\textbf{0.67} &
\cellcolor{gray!15}0.033 & \cellcolor{gray!15}\textbf{0.423} & \cellcolor{gray!15}\textbf{1.798} &
\cellcolor{gray!15}\textbf{0.07} & \cellcolor{gray!15}\textbf{0.13} & \cellcolor{gray!15}\textbf{0.25} &
\cellcolor{gray!15}\textbf{0.05} & \cellcolor{gray!15}\textbf{0.16} & \cellcolor{gray!15}\textbf{0.27} &
\cellcolor{gray!15}\textbf{0.04} & \cellcolor{gray!15}\textbf{0.16} & \cellcolor{gray!15}\textbf{0.25} \\
\bottomrule
\end{tabular}}
\end{table*}

\begin{table*}[t]
\scriptsize
\centering
\caption{Ablation study on PADG and loss function design on NAVSIM v1 and nuScenes. `Condition' denotes map and agent conditioning, while `Reference GTs' refers to reference trajectory guidance. We use ${\operatorname{TransFuser}_{\operatorname{mmt}}}^{*}$ as the baseline on NAVSIM v1 and SparseDrive as the baseline on nuScenes. \textcolor{black}{`FPS' is measured on an NVIDIA 4090 GPU.}}
\label{tab:ablation_padg_loss_nuscenes_navsim}
\renewcommand\arraystretch{0.7}
\setlength{\tabcolsep}{0.6mm}

\resizebox{\linewidth}{!}{
\begin{tabular}{cc|cc|cc| ccccccc cccc}
\toprule
\multicolumn{6}{c}{\textbf{Design / Loss}} &
\multicolumn{7}{c}{\textbf{NAVSIM v1}} &
\multicolumn{4}{c}{\textbf{nuScenes}} \\
\cmidrule(lr){1-6}\cmidrule(lr){7-13}\cmidrule(lr){14-17}
Condition & Reference GTs &
L1 Loss & $\mathcal{L}_{\text{match}}$ &
$\mathcal{L}_{\text{RL(PPO)}}$ & $\mathcal{L}_{\text{RL(GRPO)}}$ &
NC$\uparrow$ & DAC$\uparrow$ & TTC$\uparrow$ & Comf.$\uparrow$ & EP$\uparrow$ & PDMS$\uparrow$ & FPS$\uparrow$ &
$\operatorname{\textit{Div.}1s}\uparrow$ &
$\operatorname{\textit{Div.}2s}\uparrow$ &
$\operatorname{\textit{Div.}3s}\uparrow$ &
$\operatorname{\textit{Div.}Avg}\uparrow$ \\
\midrule
 &  & $\checkmark$ &  &  &  &
96.2 & 95.4 & 90.7 & 100 & 80.7 & 85.1 & \textbf{56} &
0.05 & 0.11 & 0.23 & 0.13 \\
$\checkmark$ &  & $\checkmark$ &  &  &  &
97.2 & 96.3 & 90.8 & 100 & 81.2 & 86.3 & 44 &
0.06 & 0.15 & 0.26 & 0.16 \\
 & $\checkmark$ &  & $\checkmark$ &  &  &
97.6 & 95.7 & 91.5 & 100 & 81.6 & 86.4 & 50 &
0.08 & 0.18 & 0.29 & 0.18 \\
$\checkmark$ & $\checkmark$ & $\checkmark$ &  &  &  &
96.5 & 95.5 & 90.9 & 100 & 81.0 & 85.5 & 41 &
0.06 & 0.12 & 0.23 & 0.14 \\
$\checkmark$ & $\checkmark$ & $\checkmark$ &  & $\checkmark$ &  &
96.9 & 95.5 & 91.0 & 100 & 81.2 & 85.9 & 41 &
0.06 & 0.14 & 0.25 & 0.15 \\
$\checkmark$ & $\checkmark$ &  & $\checkmark$ &  &  &
97.6 & 93.2 & 92.1 & 100 & 82.1 & 86.7 & 41 &
0.06 & 0.14 & 0.24 & 0.15 \\
$\checkmark$ & $\checkmark$ &  & $\checkmark$ & $\checkmark$ &  &
98.1 & 95.4 & 94.1 & 100 & 82.3 & 87.9 & 41 &
0.09 & \textbf{0.19} & 0.32 & 0.20 \\
\cellcolor{gray!15}$\checkmark$ & \cellcolor{gray!15}$\checkmark$ &\cellcolor{gray!15}  & \cellcolor{gray!15}$\checkmark$ & \cellcolor{gray!15} & \cellcolor{gray!15}$\checkmark$ &
\cellcolor{gray!15}\textbf{98.5} & \cellcolor{gray!15}\textbf{96.5} & \cellcolor{gray!15}\textbf{94.9} & \cellcolor{gray!15}100 & \cellcolor{gray!15}\textbf{82.6} & \cellcolor{gray!15}\textbf{88.3} & \cellcolor{gray!15}41 &
\cellcolor{gray!15}\textbf{0.10} & \cellcolor{gray!15}\textbf{0.19} & \cellcolor{gray!15}\textbf{0.34} & \cellcolor{gray!15}\textbf{0.21} \\
\bottomrule
\end{tabular}}
\end{table*}

\begin{table*}[t]
\scriptsize
\centering
\addtolength{\tabcolsep}{0.1pt}
\caption{Ablation Study on PADG on Bench2Drive. `Condition' denotes map and agent conditioning, while `Reference GTs' refers to reference trajectory guidance.
}
  \renewcommand\arraystretch{0.7}
  \tabcolsep=1.5mm 
  \resizebox{\linewidth}{!}{
  \begin{tabular}{l c cccc cccccc}
    \toprule
   \multirow{2}{*}{\textbf{$\operatorname{Method}$}}& $\operatorname{Open-loop\ Metric}$     & \multicolumn{4}{c}{\textbf{$\operatorname{Closed-loop\ Metric}$}}& \multicolumn{6}{c}{\textbf{$\operatorname{Multi-Ability (\%) \uparrow}$}}  \\
   \cmidrule(lr){2-2}\cmidrule(lr){3-6}\cmidrule(lr){7-12}
   &$\operatorname{Avg. \textit{Div.}^{(t)}}\uparrow$&$\operatorname{DS}\uparrow$ & $\operatorname{SR\ (\%)}\uparrow$ & $\operatorname{Effi}\uparrow$  & $\operatorname{Comf}\uparrow$ &
   $\operatorname{Merg.}$& $\operatorname{Overta.}$& $\operatorname{Emerge.}$& $\operatorname{Give\:Way}$ &$\operatorname{Traffic\:Sign}$& $\operatorname{Mean}$
   \\
\midrule

${\operatorname{SparseDrive}}$ & 0.21& 44.54& 16.71& 170.21& 48.63&12.18& 23.19& 17.91& 20.00& 20.98& 18.85 \\
\midrule
+ Condition & 0.24& 45.34& 18.32& 173.00& 50.34&13.21& 25.91& 18.01& 20.00& 22.87& 20.00 \\
+ Reference GTs & 0.29& 47.12& 19.89& 172.86& 51.31&14.02& 26.38& 18.78& 20.00& 22.91& 20.42 \\
\midrule
\cellcolor{gray!15}DIVER& \cellcolor{gray!15}\textbf{0.35}& \cellcolor{gray!15}\textbf{49.21}& \cellcolor{gray!15}\textbf{21.56}& \cellcolor{gray!15}\textbf{177.00}& \cellcolor{gray!15}\textbf{54.72}&\cellcolor{gray!15}\textbf{15.98}& \cellcolor{gray!15}\textbf{28.22}& \cellcolor{gray!15}\textbf{23.71} &\cellcolor{gray!15}20.00& \cellcolor{gray!15}\textbf{24.38} &\cellcolor{gray!15}\textbf{22.46} \\
\bottomrule
\end{tabular}}
\label{tab_ablation_b2d_PADG}
\end{table*}

\begin{table}[t]
\centering
  \caption{\textcolor{black}{Ablation Study on the TrajPooler on NAVSIM v1 dataset. FullImg denotes full image features without projection and NoImg denotes no image features (trajectory encoding only).}}
\renewcommand\arraystretch{0.7}
  \tabcolsep=1.5mm 
  \resizebox{\linewidth}{!}{
\begin{tabular}{l  cccccc}
\toprule
\multicolumn{1}{l}{Method}  & NC$\uparrow$                & \multicolumn{1}{c}{DAC}$\uparrow$ & TTC$\uparrow$                 & Comf.$\uparrow$                & EP$\uparrow$                  & PDMS$\uparrow$ \\
\midrule

${\operatorname{TransFuser}_{\operatorname{mmt}}}^{*}$&96.2& 95.4& 90.7& 100& 80.7& 85.1 \\ 

\midrule
DIVER+NoImg    &  96.9 & 95.6 & 94.1 & 100 & 81.5 & 87.7  \\ 
DIVER+FullImg  &  97.3 & 95.9 & 94.5  &100 & 81.9 & 87.9  \\ 
\midrule
\cellcolor{gray!15}DIVER+TrajPooler  &  \cellcolor{gray!15}\textbf{98.5} &\cellcolor{gray!15}\textbf{96.5}& \cellcolor{gray!15}\textbf{94.9} &\cellcolor{gray!15}100 & \cellcolor{gray!15}\textbf{82.6}& \cellcolor{gray!15}\textbf{88.3} \\ 
\bottomrule
\end{tabular}}
\label{tab_Ablation_navsim_trajpooler}
\end{table}

\begin{table*}[t]
\scriptsize
\centering
\renewcommand\arraystretch{0.7}
\setlength{\tabcolsep}{0.25mm}
\caption{\textcolor{black}{Ablation study on different reward components on Bench2Drive, nuScenes, and NAVSIM v1. We use SparseDrive as the baseline for Bench2Drive/nuScenes and ${\operatorname{TransFuser}_{\operatorname{mmt}}}^{*}$ as the baseline for NAVSIM v1.}}
\label{tab:ablation_reward}

\resizebox{\linewidth}{!}{
\begin{tabular}{ccccc ccccc ccc ccc cccccc}
\toprule
\multicolumn{5}{c}{\textbf{$\operatorname{Reward}$}} &
\multicolumn{5}{c}{\textbf{Bench2Drive}} &
\multicolumn{6}{c}{\textbf{nuScenes}} &
\multicolumn{6}{c}{\textbf{NAVSIM v1}} \\
\cmidrule(lr){1-5}\cmidrule(lr){6-10}\cmidrule(lr){11-16}\cmidrule(lr){17-22}
$\lambda_{\text{div}}$ & $\lambda_{\text{safe}}$ & $\lambda_{\text{TC}}$ & $\lambda_{\text{LK}}$ & $\lambda_{\text{PDMS}}$ &
$\operatorname{Avg.\ \textit{Div.}^{(t)}}\uparrow$ &
$\operatorname{DS}\uparrow$ &
$\operatorname{SR\ (\%)}\uparrow$ &
$\operatorname{Effi}\uparrow$ &
$\operatorname{Comf}\uparrow$ &
$\operatorname{\textit{Div.}1s}\uparrow$&
$\operatorname{\textit{Div.}2s}\uparrow$&
$\operatorname{\textit{Div.}3s}\uparrow$&
$\operatorname{Col.1s}\uparrow$&
$\operatorname{Col.2s}\uparrow$&
$\operatorname{Col.3s}\uparrow$&

NC$\uparrow$ & DAC$\uparrow$ & TTC$\uparrow$ & Comf.$\uparrow$ & EP$\uparrow$ & PDMS$\uparrow$ \\
\midrule
 &  &  &  &  &
0.21 & 44.54 & 16.71 & 170.21 & 48.63 &
0.05 & 0.11 & 0.23 & 
0.01 & 0.05 & 0.18 & 
96.2 & 95.4 & 90.7 & 100 & 80.7 & 85.1 \\

0.5 & 0.3 &  &  &  &
0.35 & 49.21 & 21.56 & 177.00 & 54.72 &
0.10 & 0.19 & 0.34 & 
0.01 & 0.05 & 0.15 & 
98.5 & 96.5 & 94.9 & 100 & 82.6 & 88.3 \\

0.5 & 0.3 & 0.1 &  &  &
0.33 & 48.71 & 21.40 & 176.81 & 56.10 &
0.10 & 0.18 & 0.35 & 
0.01 & 0.05 & 0.16 & 
-- & -- & -- & -- & -- & -- \\

0.5 & 0.3 & 0.1 & 0.1 &  &
0.32 & 48.89 & 21.37 & 176.34 & 56.23 &
0.09 & 0.17 & 0.35 & 
0.01 & 0.05 & 0.16 & 
-- & -- & -- & -- & -- & -- \\

0.5 & 0.3 &  &  & 0.1 &
-- & -- & -- & -- & -- &
-- & -- & -- & 
-- & -- & -- & 
98.9 & 96.6 & 94.7 & 100 & 82.3 & 88.6 \\

\bottomrule
\end{tabular}}
\end{table*}

To demonstrate the generalization capability of \textbf{DIVER}, we conduct experiments against several strong E2E-AD baselines, including \textbf{SparseDrive}~ \cite{sun2024sparsedrive}, \textbf{VAD}~ \cite{jiang2023vad}, and \textbf{TransFuser} \cite{TransFuser}.
On the \textbf{Bench2Drive}~ \cite{jia2024bench2drive}, \textbf{nuScenes}~ \cite{nuscenes}, and \textbf{Adv-nuSc}~ \cite{xu2025challenger} datasets, we compare against \textbf{SparseDrive} and \textbf{VAD}. For nuScenes and Adv-nuSc, we adopt ResNet-50~ \cite{resnet} as the image backbone with an input resolution of 256$\times$704. The detection module operates within a circular range of 55 meters, and the online mapping covers a 60$\times$30 meter area (longitudinal$\times$lateral). The motion module generates 6 trajectory candidates (modes).
On \textbf{Bench2Drive}, we use ResNet-50 with 6 decoder layers and an input resolution of 640$\times$352. We also define fixed numbers of hybrid task queries: 900 agent queries, 100 map queries, and 480 planning queries.
On the \textbf{NAVSIM}~ \cite{dauner2024navsim} dataset, we adopt \textbf{TransFuser}~ \cite{TransFuser} as the baseline and follow the standard Navtrain split for training. For fair comparison, we use the same perception modules and ResNet-34 backbone as in TransFuser.
All experiments are conducted on a server with 8 NVIDIA RTX 4090 (24GB) GPUs. \textcolor{black}{On the NuScenes dataset, the total batch size is set to 48, with 20 training epochs, resulting in an overall training time of 11.39 hours.
On the Bench2Drive dataset, the total batch size is also 48, with 2 training epochs, and the overall training time is 51.38 hours.
On the NAVSIM v1 dataset, the total batch size is 64, with 100 training epochs, and the overall training time is 3.39 hours.}

\subsection{Main Results}
To comprehensively evaluate the multi-mode trajectory generation capability of our proposed \textbf{DIVER}, we conduct \textbf{closed-loop} experiments on three representative E2E-AD benchmarks: \textbf{Bench2Drive},  \textbf{NAVSIM} and \textbf{NuScenes}, as shown in Tables \ref{tab_b2d}, \ref{tab_navsimv1},\ref{tab_navsimv2} and \ref{tab_nuscenes_planning}. It is important to note that we do \underline{not} adopt the conventional \textbf{L2 distance} in open-loop evaluation. This is because L2 distance merely reflects the proximity between a predicted trajectory and a single GT trajectory, failing to capture the diversity and spread of multi-mode trajectory predictions. Relying solely on this metric compromises the core objective of multi-mode trajectory generation—namely, to cover a wide range of plausible future intents.

\subsubsection{Close-Loop Results}
\textbf{Bench2Drive (Close-Loop).}
\textcolor{black}{In Table \ref{tab_b2d}, DIVER significantly outperforms VAD, ${\operatorname{VAD}_{\operatorname{mmt}}}$, MomAD, and SparseDrive across all planning metrics. Compared to ${\operatorname{VAD}_{\operatorname{mmt}}}$, it improves Success Rate by 11.8\%, Comfortness by 8.6\%, Driving Score by 4.1\%, and Efficiency by 4.0\%. Against SparseDrive, DIVER achieves even larger gains: +29.0\% in Success Rate, +12.5\% in Comfortness, +10.5\% in Driving Score, and +4.0\% in Efficiency.
Compared to DiffAD, our DIVER achieves a higher Driving Score (68.90 vs. 67.92) and multi-ability mean (42.09 vs. 38.79), with clear gains on Merg. (+5.08\%), Give Way (+10.00\%), and Traffic Sign (+12.89\%), while DiffAD attains a slightly higher Success Rate (38.64\% vs. 36.75\%). Moreover, in multi-ability evaluation, DIVER achieves the highest average score, with notable gains in overtaking (+3.78\%) and emergency braking (+5.51\%), reflecting improved decision-making and semantic awareness. This demonstrates that DIVER not only improves planning robustness but also enables flexible driving behaviors.
}

\noindent\textbf{NAVSIM v1 (Closed-Loop).}
Table~\ref{tab_navsimv1} shows that \textbf{DIVER} achieves strong closed-loop performance on \textbf{NAVSIM v1} while maintaining full comfort (100). With C\&L inputs, DIVER improves PDMS to 88.3, outperforming TransFuser and ${\operatorname{TransFuser}_{\operatorname{mmt}}}^{*}$, and remains competitive with recent diffusion- and RL-based planners. These gains are mainly driven by improved DAC and TTC, indicating safer and more compliant interactions.
Compared with Hydra-MDP \cite{li2024hydra}, which distills PDMS metrics into a static policy in a single-step process, DIVER dynamically incorporates PDMS-guided reinforcement within the diffusion-based trajectory generation. These results highlight DIVER's ability to balance safety, comfort, and goal-directed planning, demonstrating its effectiveness in realistic driving simulations. The improvements across all aspects suggest that our method not only generates diverse multi-mode trajectories but also ensures their practical executability and robustness in closed-loop scenarios.

\noindent\textbf{NAVSIM v2 (Closed-Loop).}
As shown in Tables~\ref{tab_navsimv2} and~\ref{tab_navsimv2_navhard}, DIVER achieves competitive closed-loop performance on NAVSIM v2. On the navtest split, it attains a strong EPDMS with top-tier traffic-light compliance and hard-constraint safety, indicating reliable rule compliance. On the more challenging navhard split, DIVER achieves an EPDMS of \textbf{26.8}, outperforming LTF and DiffusionDrive and remaining close to GuideFlow, while maintaining robust performance under the perturbed Stage~2 evaluation, suggesting improved resilience to distribution shifts.
\begin{figure}[htp]
\centering
 \includegraphics[width=1.0\linewidth]{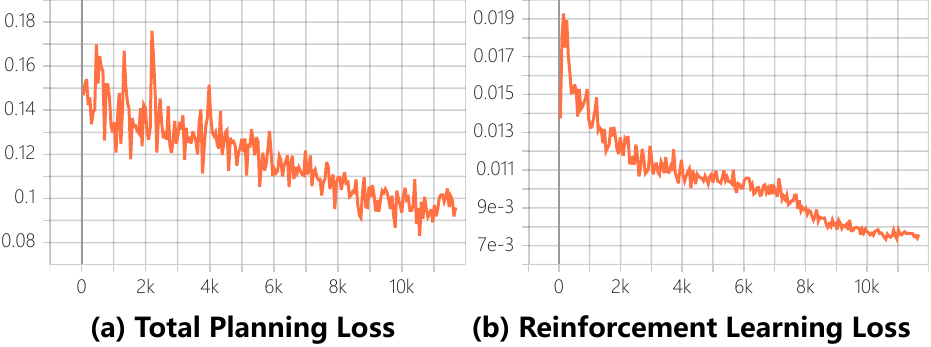}
\caption[ ]{\textcolor{black}{Visualization of Planning Loss and Reinforcement Learning Loss of DIVER on the nuScenes \cite{nuscenes} dataset, showing stable convergence without oscillation or divergence.}}
\label{fig:vis_loss}
\end{figure}

\begin{figure}[htp]
\centering
 \includegraphics[width=1.0\linewidth]{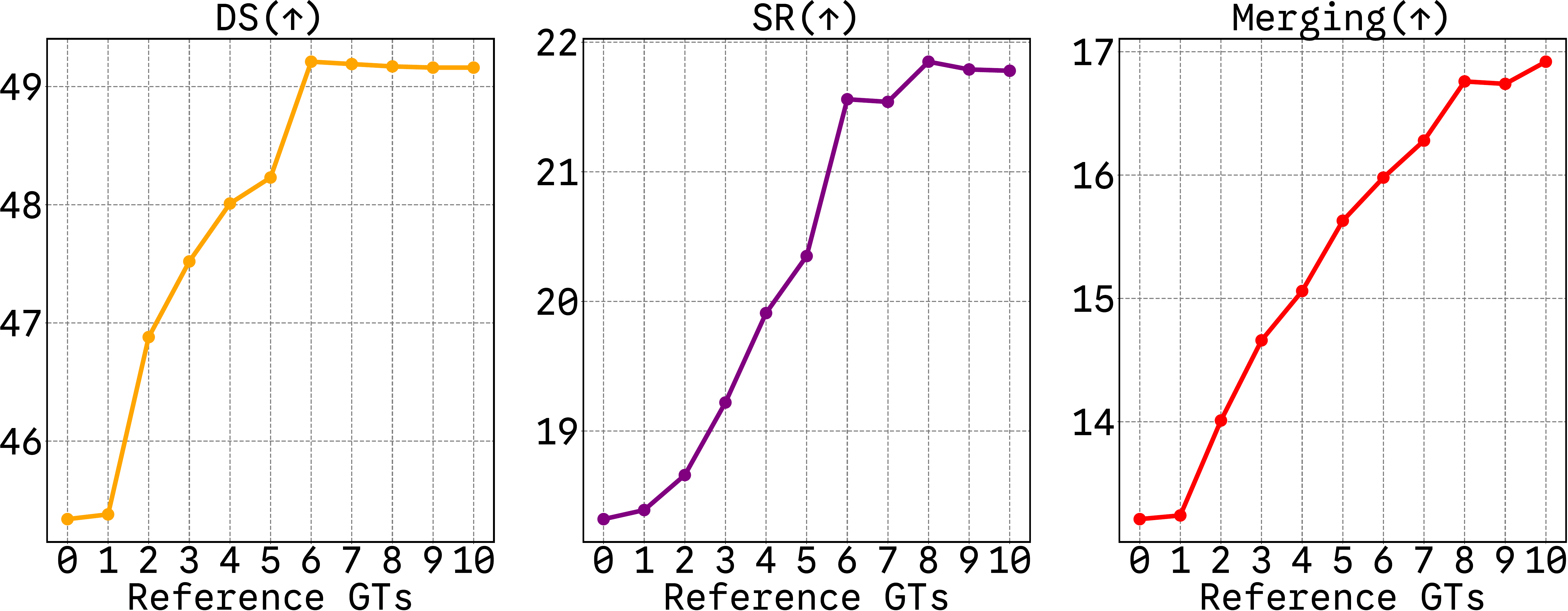}
\caption[ ]{\textcolor{black}{Impact of the Number of Reference GTs on Closed-Loop Performance (Bench2Drive). 
A value of 0 indicates no reference GTs (only the GT trajectory is used). Values of 1 and above correspond to using one reference GTs in addition to the GT, and so on.}}
\label{fig:Ablation_ReferenceGTs_PADG}
\end{figure}

\begin{figure}[htp]
\centering
 \includegraphics[width=1.0\linewidth]{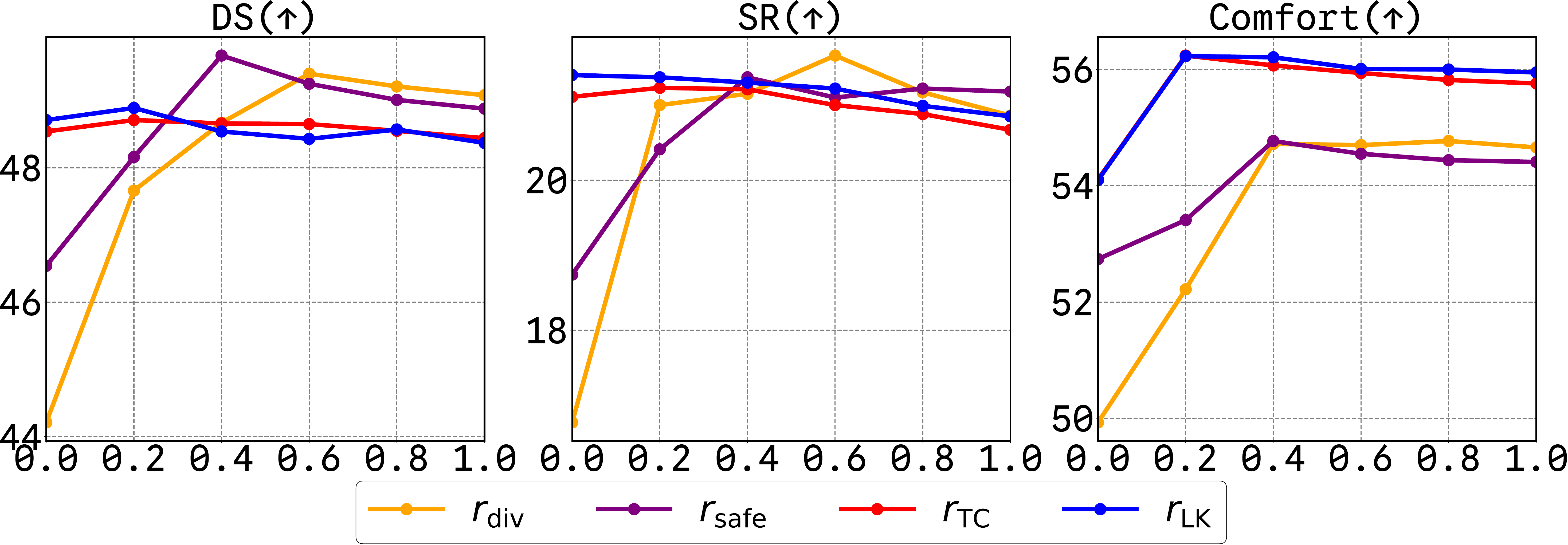}
\caption[ ]{\textcolor{black}{Impact of reward-weight sensitivity on closed-loop performance (Bench2Drive).Each curve sweeps one weight $\lambda$ with the others fixed at $(0.5,0.3,0.1,0.1)$,
except when sweeping $\lambda_{\text{div}}$ (fixing $\lambda_{\text{safe}}=0.3$) or
$\lambda_{\text{safe}}$ (fixing $\lambda_{\text{div}}=0.5$).
}}
\label{fig:Ablation_reward_weight_sensitivity}
\end{figure}

\begin{figure*}[htp]
\centering
 \includegraphics[width=1.0\linewidth]{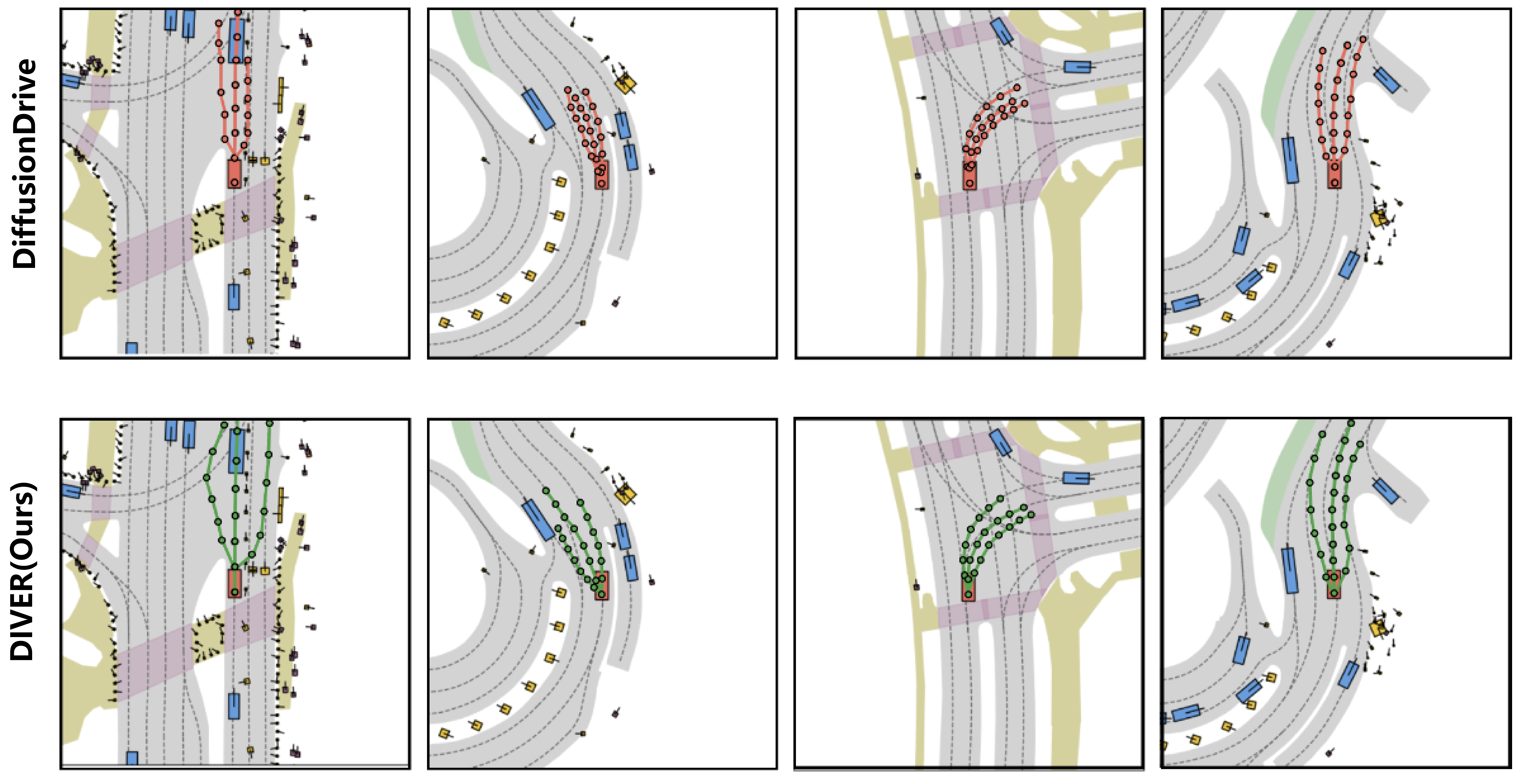}
\caption[ ]{\textcolor{black}{Visualization results of DIVER compared with DiffusionDrive  \cite{liao2024diffusiondrive} on the NAVSIM  \cite{dauner2024navsim} dataset. Our proposed DIVER generates significantly more diverse and feasible trajectories across various driving scenarios.}}
\label{fig:vis_navsim}
\end{figure*}

\subsubsection{Open-Loop Results}
\noindent\textbf{NuScenes (Open-Loop).}
As shown in Table \ref{tab_nuscenes_planning}, DIVER demonstrates superior performance on the nuScenes dataset in both diversity and safety. For multi-mode trajectory generation, it achieves Diversity Metric scores of 0.10, 0.19, 0.34 at 1s, 2s, and 3s, with an average of 0.21—representing a 61.5\% improvement over DiffusionDrive (avg. 0.15). This indicates more intention-aware and temporally balanced trajectory generation. In terms of safety, DIVER reduces the average collision rate to 0.07, compared to 0.08 from DiffusionDrive, achieving a 12.5\% reduction. Notably, this improvement in diversity does not come at the cost of safety. These results confirm the effectiveness of DIVER in producing diverse and safe trajectories.

\noindent\textbf{Bench2Drive (Open-Loop).}
As shown in Table \ref{tab_b2d}, DIVER achieves significantly higher diversity in Open-Loop evaluation on the Bench2Drive dataset. It reaches a score of 0.32 compared to 0.20 for $\text{VAD}_{\text{mmt}}$ and 0.18 for MomAD, marking a 60.0\% improvement. Against SparseDrive (0.21), DIVER scores 0.35, improving by 66.7\%. These results highlight DIVER’s stronger ability to generate diverse and intention-rich trajectories.

\subsection{Robustness Study}
\textbf{Long-Horizon Planning (NuScenes).}
In Table~\ref{tab:robustness_study}, we evaluate long-horizon planning on nuScenes with a 6-second horizon. DIVER achieves the highest Diversity Metric at all timestamps (0.50/0.61/0.75 at 4s/5s/6s; avg. 0.62), outperforming SparseDrive (avg. 0.43) and DiffusionDrive (avg. 0.49) by 44.2\% and 26.5\%, respectively. Meanwhile, DIVER yields the lowest collision rate at 6s (1.91\%), improving over DiffusionDrive (2.19\%) and MomAD (2.33\%). Overall, DIVER preserves multi-mode diversity over long horizons while maintaining strong safety.

\noindent\textbf{Turning-Heavy Scenarios (Turning-nuScenes).}
Table~\ref{tab:robustness_study} shows that DIVER is more robust in turning-heavy scenarios. It achieves higher diversity (avg. 0.31) than DiffusionDrive (avg. 0.20) and MomAD (avg. 0.19), indicating richer trajectory modes. DIVER also attains the lowest collision rate across horizons (avg. 0.11), halving MomAD’s average (0.22). These results confirm improved robustness under complex multi-turn maneuvers.

\noindent\textbf{Adversarial Perturbation Settings (Adv-nuSc).}
On Adv-nuSc (Table~\ref{tab:robustness_study}), DIVER achieves the lowest collision rates at all horizons (avg. 0.752\%), outperforming UniAD (3.95\%), VAD (7.05\%), SparseDrive (2.06\%), and DiffusionDrive (1.67\%). This demonstrates better safety robustness under worst-case conditions.

\noindent\textbf{Robustness in Noisy-Weather Scenarios (nuScenes-C).}
As shown in Table~\ref{tab:robustness_study}, DIVER consistently achieves the lowest collision rates across weather corruptions. Under clean scenes, DIVER matches the best baseline (0.05\%). Under snow/rain/fog, DIVER remains more robust with average collision rates of 0.09\%/0.15\%/0.21\%, outperforming SparseDrive (0.17\%/0.35\%/0.32\%) and DiffusionDrive (0.13\%/0.18\%/0.26\%). These results indicate better generalization and safer planning under degraded sensing conditions.

\begin{figure*}[t]
\centering
 \includegraphics[width=1.0\linewidth]{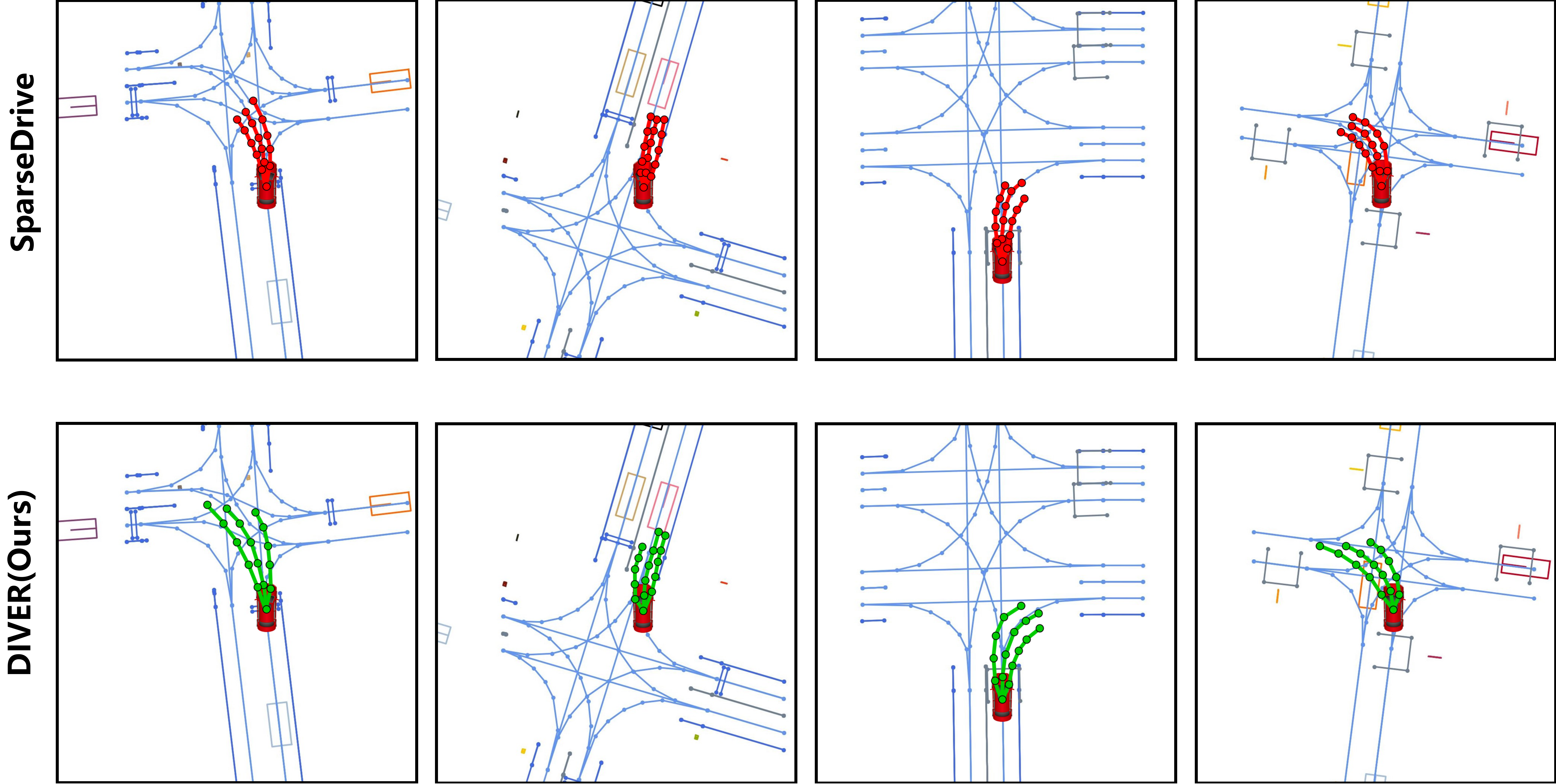}
\caption[ ]{\textcolor{black}{Visualization results of DIVER compared with SparseDrive  \cite{sun2024sparsedrive} on the Bench2Drive  \cite{jia2024bench2drive} dataset. It further confirms DIVER's powerful capability in generating diverse trajectories.}}
\label{fig:vis_b2d}
\end{figure*}

\begin{figure}[htp]
\centering
 \includegraphics[width=1.0\linewidth]{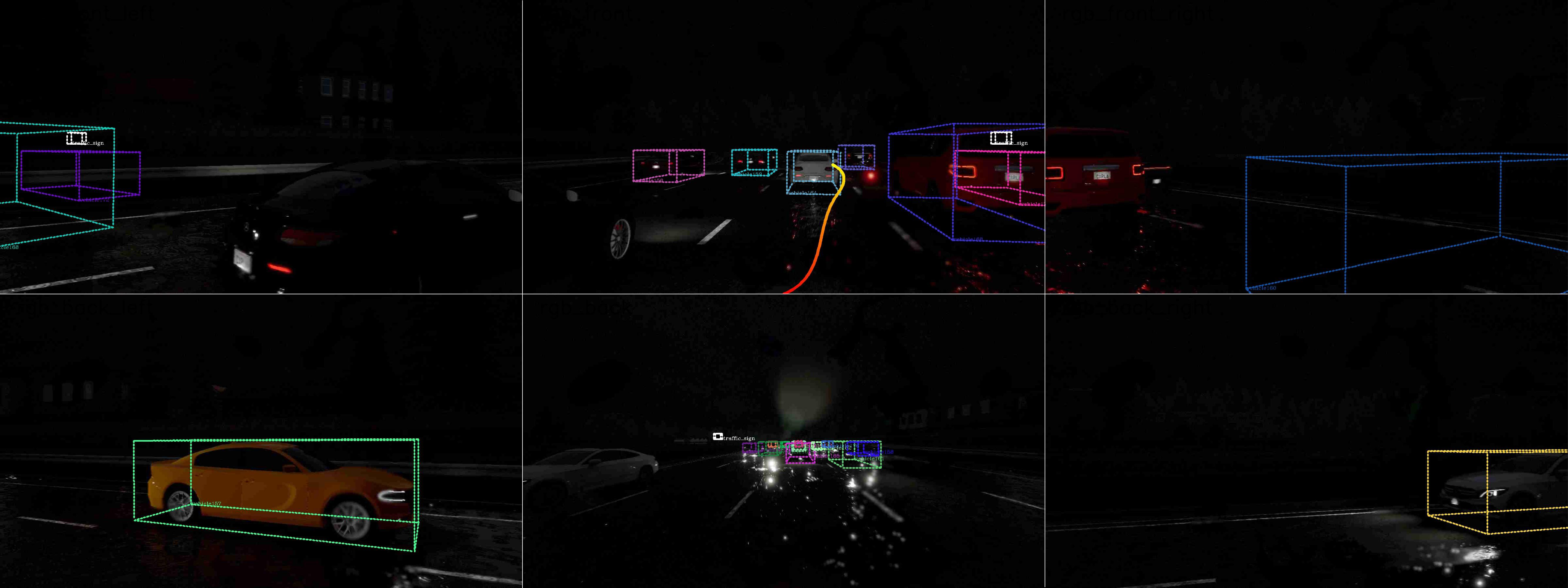}
\caption[ ]{\textcolor{black}{Visualization of representative failure cases of DIVER on Bench2Drive, focusing on long-tail conditions such as nighttime driving, where reduced visibility and rare interactions can still lead to suboptimal planning behaviors.}}
\label{fig:vis_b2d_failure_case}
\end{figure}

\subsection{Ablation Study}
\noindent\textbf{Ablation Study on PADG and Loss Function Design.}
\textcolor{black}{As shown in Tables~\ref{tab:ablation_padg_loss_nuscenes_navsim} and~\ref{tab_ablation_b2d_PADG}, we conduct comprehensive ablations on  NAVSIM v1~\cite{dauner2024navsim}, nuScenes~\cite{nuscenes}, and Bench2Drive~\cite{jia2024bench2drive}, to validate the key design choices in PADG.
Where, Table~\ref{tab:ablation_padg_loss_nuscenes_navsim} shows that loss design is crucial for fully exploiting PADG. On nuScenes, the standard L1 objective provides limited improvements in diversity, whereas replacing it with the proposed matching loss $\mathcal{L}_{\text{match}}$ enables more effective utilization of reference guidance and leads to consistently higher $\operatorname{\textit{Div.}^{(t)}}$.
Incorporating reinforcement learning further improves planning quality, and GRPO yields stronger and more stable gains than PPO, achieving the best trade-off between safety and performance on NAVSIM v1 (e.g., top PDMS while maintaining competitive NC/DAC/TTC).
Finally, the training curves in Fig.~\ref{fig:vis_loss} confirm stable optimization, where the planning and RL losses converge smoothly without oscillation, supporting the effectiveness of the proposed PADG and loss design.
On Bench2Drive, adding \emph{Condition} consistently improves both trajectory diversity and closed-loop planning quality over SparseDrive, showing the benefit of context-aware diffusion. Incorporating \emph{Reference GTs} further mitigates mode collapse and brings additional gains in DS/SR and multi-ability. Combining both (DIVER) achieves the best overall results, highlighting their complementary effects.}


\noindent\textbf{Impact of the Number of Reference GTs.} As shown in Figure \ref{fig:Ablation_ReferenceGTs_PADG}, increasing the number of Reference GTs leads to consistent gains in Driving Score, Success Rate, and Merging. Performance stabilizes around 6–7 GTs, indicating that moderate reference diversity effectively enhances planning robustness and behavioral richness.

\noindent\textbf{Impact of the TrajPooler module.}
\textcolor{black}{TrajPooler is designed to focus on image regions that are spatially correlated with the predicted trajectories rather than processing the entire image.
As shown in Table \ref{tab_Ablation_navsim_trajpooler}, to evaluate the effectiveness of the proposed TrajPooler module, we conduct ablation experiments comparing three configurations: full image features without projection, no image features (trajectory encoding only), and the TrajPooler module.
These results indicate that TrajPooler effectively aggregates perception features around the predicted trajectory, leading to a 1.2\% increase in NC and a 0.4\% increase in PDMS compared with using full image features, thereby enabling the model to focus on spatially relevant cues while avoiding redundant global contexts.
The results demonstrate that TrajPooler not only reduces computational cost but also enhances the model’s ability to focus on trajectory-relevant regions, leading to improved representation efficiency.}


\noindent\textbf{Impact of Different Reward Mechanisms.}
As shown in Table \ref{tab:ablation_reward}, incorporating $r_{\text{div}}$ and $r_{\text{safe}}$ leads to notable quantitative improvements across all datasets. On Bench2Drive, the average diversity $\operatorname{Div.}^{(t)}$ increases from 0.21 to 0.35 (+66.7\%), and the success rate rises from 16.71\% to 21.56\% (+29.1\%). On NuScenes, the average trajectory diversity improves from 0.13 to 0.21 (+61.5\%) while maintaining a low collision rate (0.08→0.07). On NAVSIM v1, combining $r_{\text{div}}$, $r_{\text{safe}}$, and PDMS guidance yields the best performance, with NC improving from 96.2 to 98.9 (+2.7\%) and DAC from 95.4 to 96.6 (+1.3\%). These quantitative results confirm that multi-term reward design significantly enhances both trajectory diversity and driving safety across benchmarks.
\textcolor{black}{
As shown in Fig.~\ref{fig:Ablation_reward_weight_sensitivity}, we study the sensitivity of closed-loop performance to the reward weights $\lambda$. We observe that $\lambda_{\text{div}}$ and $\lambda_{\text{safe}}$ have the most pronounced impact: performance improves as their weights increase and then gradually saturates beyond a certain range. In contrast, $\lambda_{\text{TC}}$ and $\lambda_{\text{LK}}$ mainly affect driving smoothness, highlighting their importance for comfort performance.}

\subsection{Visualization}
\noindent\textbf{Visualization of Planning Behaviors.}
As shown in Figures \ref{fig:vis_navsim} and \ref{fig:vis_b2d}, we present a qualitative comparison among DiffusionDrive, SparseDrive, and our proposed DIVER in representative turning scenarios. DIVER generates a broader spread of trajectories, effectively capturing diverse driving intentions. This demonstrates its capability to overcome the limitations of imitation learning, which often results in conservative and mode-collapsed behaviors.

\noindent\textbf{Visualization Failure Case of DIVER.} \textcolor{black}{As shown in Figure~\ref{fig:vis_b2d_failure_case}, we visualize typical long-tail scenario (e.g., nighttime driving) where DIVER may fail, highlighting the remaining challenges under low-visibility and rare conditions. 
These failures suggest that richer uncertainty-aware modeling and stronger long-tail supervision (e.g., targeted data augmentation or curriculum training on rare scenarios) are needed to further improve robustness in such safety-critical settings.}

\section{Conclusion}\label{sec:conclusion}
The proposed DIVER framework addresses a key challenge in E2E-AD: the planning model collapse commonly observed in imitation learning. By effectively integrating diffusion models with reinforcement learning, DIVER enables the generation of diverse trajectories. Compared with state-of-the-art approaches, our method demonstrates superior performance in both diversity and safety across multiple benchmarks such as Bench2Drive, NAVSIM, and nuScenes, highlighting its potential for robust and flexible autonomous planning in complex real-world scenarios. While the proposed DIVER emphasizes trajectory diversity, ensuring the correctness of trajectory selection remains a challenge. In future work, we plan to integrate the Vision-Language-Action model to jointly optimize trajectory generation and decision-making for safer planning.

\section*{ACKNOWLEDGMENTS}
This work was supported by the National Natural Science Foundation of China (NSFC) under Grants No. 62536001 (Key Program) and No. 62576026.

\ifCLASSOPTIONcaptionsoff
  \newpage
\fi

\bibliographystyle{IEEEtran}
\bibliography{egbib}
\begin{IEEEbiography}[{\includegraphics[width=1in,height=1.25in,clip,keepaspectratio]{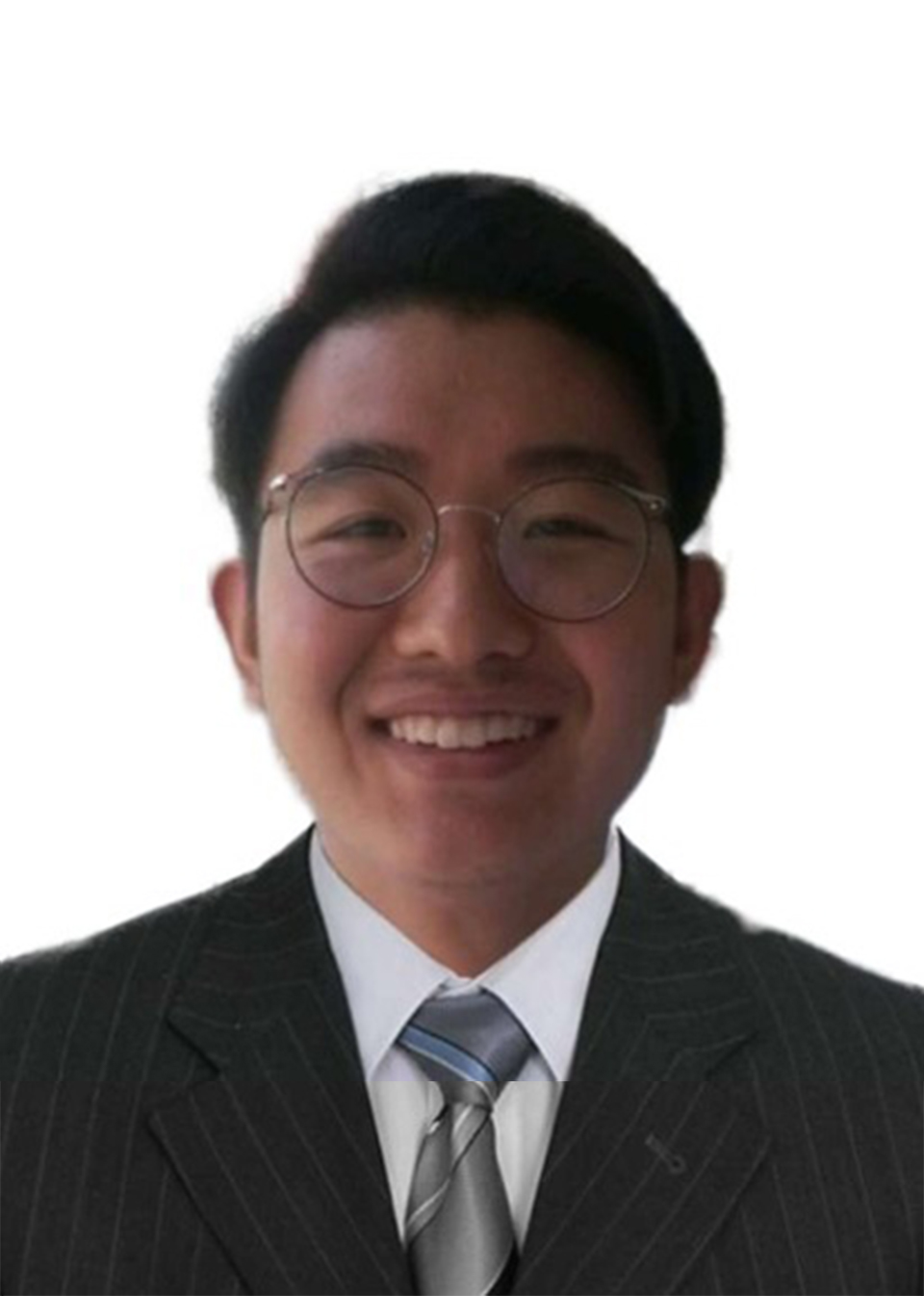}}]{Ziying Song}
 received the B.S. degree from Hebei Normal University of Science and Technology, China, in 2019, the M.S. degree from Hebei University of Science and Technology, China, in 2022, and the Ph.D. degree in computer science and technology from Beijing Jiaotong University, China, in March 2026. He is currently an Assistant Professor with the School of Artificial Intelligence, Yanshan University, China. His research interests include  autonomous driving, robust perception, end-to-end autonomous driving, world models, embodied intelligence, and vision-language-action models.
\end{IEEEbiography} \vspace{-2em}
\begin{IEEEbiography}
[{\includegraphics[width=1in,height=1.25in,clip,keepaspectratio]{{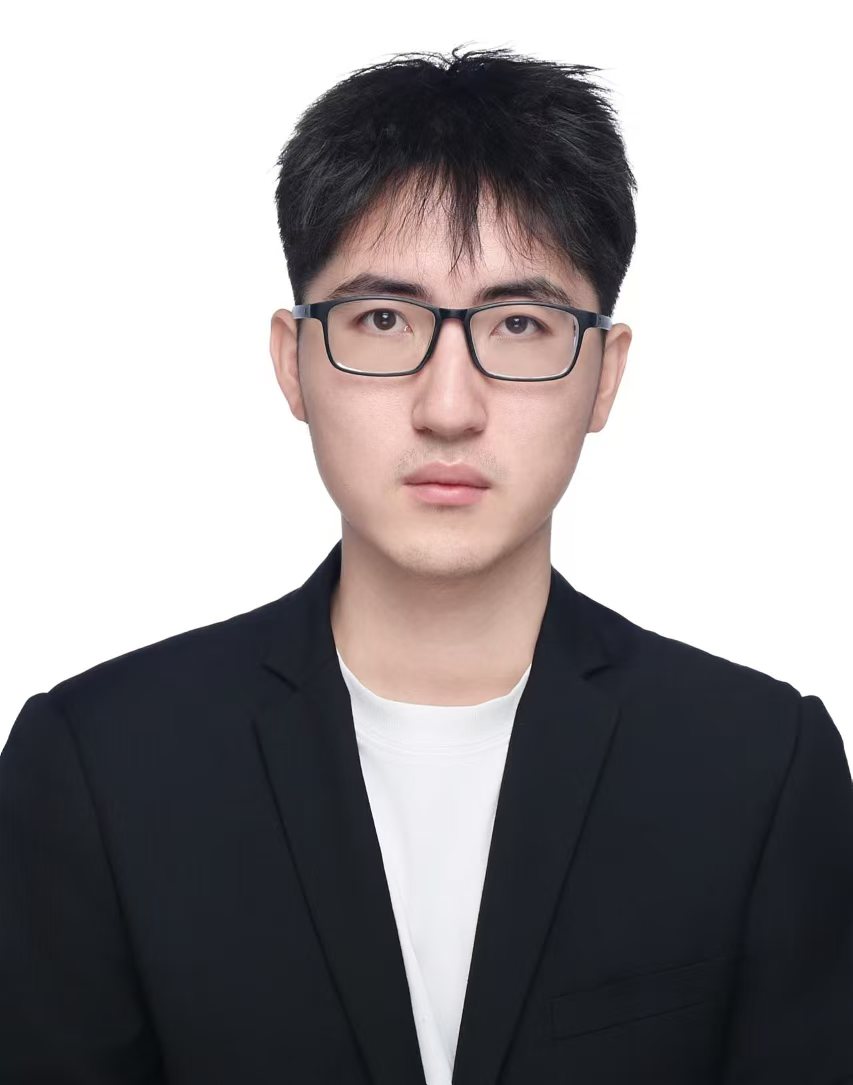}}}] {Lin Liu} was born in Jinzhou, Liaoning Province, China, in 2001. He is now a college student majoring in Computer Science and Technology at China University of Geosciences(Beijing).
Since Dec. 2022, he has been recommended for a master's degree in Computer Science and Technology at Beijing Jiaotong University. His research interests are in computer vision.
\end{IEEEbiography} \vspace{-2em}
\begin{IEEEbiography}[{\includegraphics[width=1in,height=1.25in,clip,keepaspectratio]{{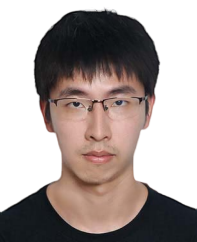}}}]{Hongyu Pan} received the B.E. degree from Beijing Institute of Technology (BIT) in 2016 and the M.S. degree in computer science from the Institute of Computing Technology (ICT), University of Chinese Academy of Sciences (UCAS), in 2019. He is currently an employee at Horizon Robotics. His research interests include computer vision, pattern recognition, and image processing. He specifically focuses on 3D detection/segmentation/motion and depth estimation.

\end{IEEEbiography} \vspace{-2em}

\begin{IEEEbiography}[{\includegraphics[width=1in,height=1.25in,clip,keepaspectratio]{{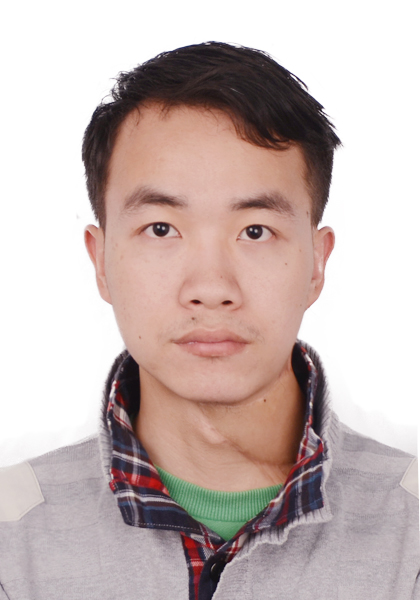}}}]{Bencheng Liao} received the B.E. degree from School of Electronic Information and Communications, Huazhong University of Science and Technology, Wuhan, China, in 2020. He is currently a PhD candidate at the Institute of Artificial Intelligence and School of Electronic Information and Communications, Huazhong University of Science and Technology. His research interests include object detection, 3D vision, and autonomous driving.
\end{IEEEbiography} \vspace{-2em}

\begin{IEEEbiography}[{\includegraphics[width=1in,height=1.25in,clip,keepaspectratio]{{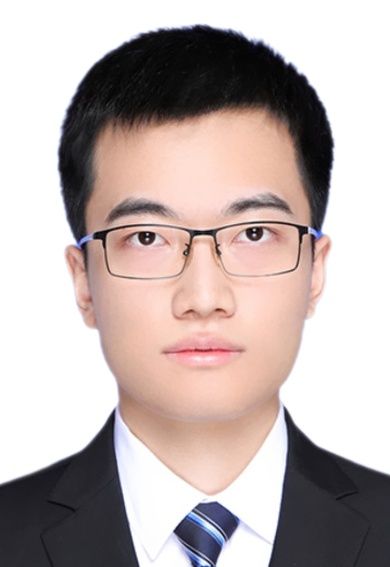}}}]{Mingzhe Guo}
is a Ph.D. candidate working on computer vision and deep learning at Beijing Key Lab of Traffic Data Analysis and Mining, Beijing Jiaotong University as of 2020. Before that he received the B.S. degree at Beijing Jiaotong University in 2020. He is about to work as a researcher at Baidu Inc. His research interest includes visual object tracking, BEV detection/tracking and E2E-AD.
\end{IEEEbiography} \vspace{-2em}

\begin{IEEEbiography}[{\includegraphics[width=1in,height=1.25in,clip,keepaspectratio]{{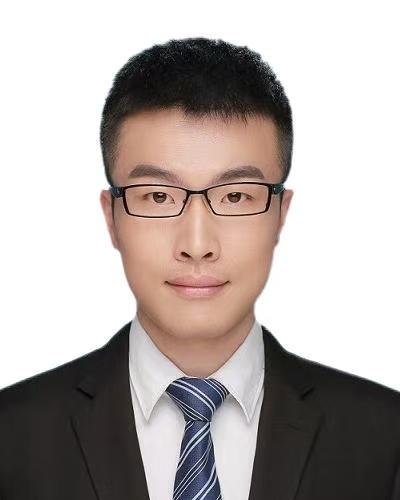}}}]
{Lei Yang (Member, IEEE)} received his M.S. degree from the Robotics Institute at Beihang University, in 2018. and the Ph.D. degree from the School of Vehicle and Mobility, Tsinghua University, in 2024. From 2018 to 2020, he joined the Autonomous Driving R\&D Department of JD.COM as an algorithm researcher. Currently, he is a research fellow with the School of Mechanical and Aerospace Engineering, Nanyang Technological University, Singapore. His current research interests include autonomous driving, 3D scene understanding and world model.
\end{IEEEbiography} \vspace{-2em}

\begin{IEEEbiography}[{\includegraphics[width=1in,height=1.25in,clip,keepaspectratio]{{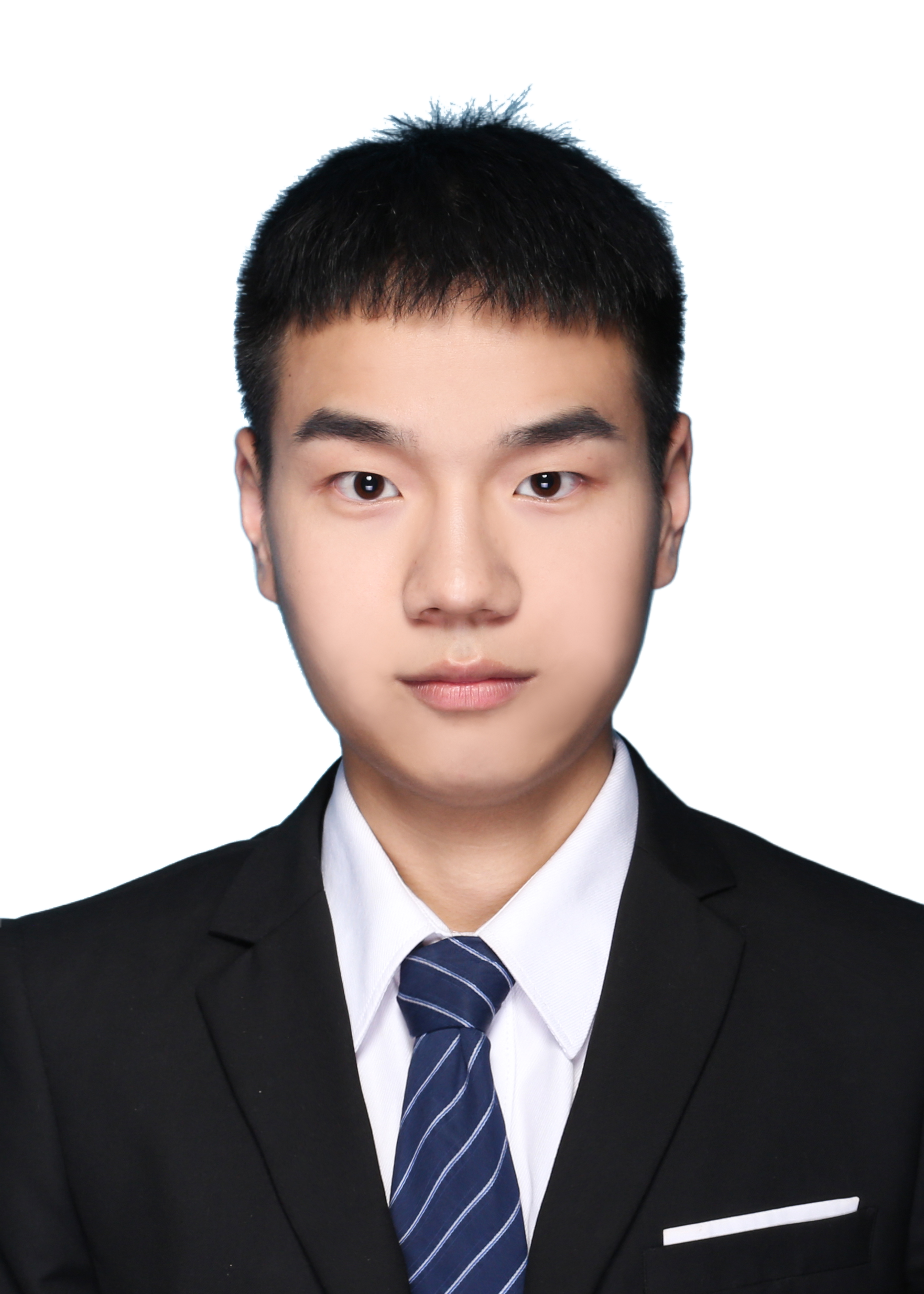}}}]{Yongchang Zhang} received the B.E. degree from the University of Electronic Science and Technology of China (UESTC) in 2020 and the M.S. degree in control theory and control engineering from the Institute of Automation (CASIA), University of Chinese Academy of Sciences (UCAS) in 2023. He is currently a Research Engineer at Horizon Robotics. His research interests include computer vision, autonomous driving perception, and visual-language models. \end{IEEEbiography} \vspace{-2em}

\begin{IEEEbiography}[{\includegraphics[width=1in,height=1.25in,clip,keepaspectratio]{{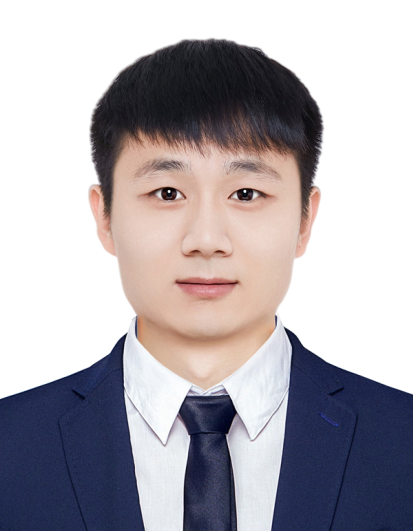}}}]{Shaoqing Xu} received his M.S. degree in transportation engineering from the School of Transportation Science and Engineering in Beihang University. He is currently working toward the Ph.D. degree in electromechanical engineering with the State Key Laboratory of Internet of Things for Smart City, University of Macau, Macao SAR, China. His research interests include 3D Space Intelligence, End2End, WorldModel, Vision-Language-Action(VLA), and its applications in Autonomous Driving and Robotics.
\end{IEEEbiography} \vspace{-2em}

\begin{IEEEbiography}[{\includegraphics[width=1in,height=1.25in,clip,keepaspectratio]{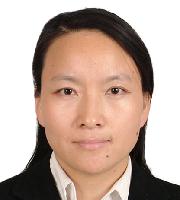}}]{Caiyan Jia}
Caiyan Jia received the B.S. degree in mathematics from Ningxia University in 1998, the M.S. degree in computational mathematics from Xiangtan University in 2001, and the Ph.D. degree in engineering from the Institute of Computing Technology, Chinese Academy of Sciences, in 2004. She is currently a Professor with the School of Computer Science and Technology, Beijing Jiaotong University, Beijing, China.
\end{IEEEbiography} \vspace{-2em}

\begin{IEEEbiography}[{\includegraphics[width=1in,height=1.25in,clip,keepaspectratio]{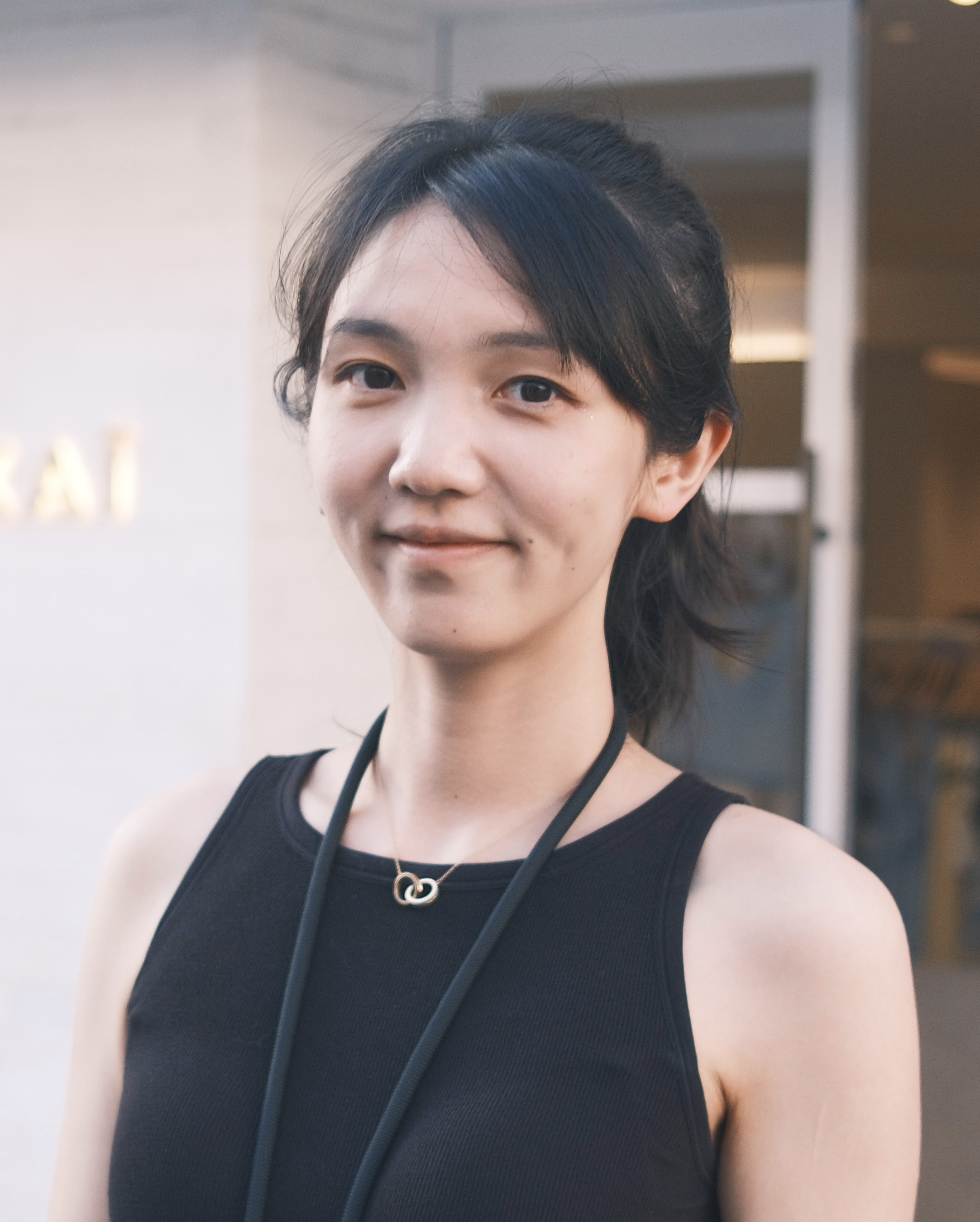}}]{Yadan Luo}(Member, IEEE) received the BS degree in computer science from the University of Electronic Science and Technology of China, and the PhD degree from the University of Queensland. Her research interests include machine learning, computer vision, and multimedia data analysis. She is now a senior lecturer and ARC DECRA at the University of Queensland.
\end{IEEEbiography} \vspace{-2em}

\end{document}

%% file: shortcuts.tex
\makeatletter
\DeclareRobustCommand\onedot{\futurelet\@let@token\@onedot}
\def\@onedot{\ifx\@let@token.\else.\null\fi\xspace}

\makeatother

\definecolor{darkgreen}{rgb}{0,0.7,0}
\definecolor{darkblue}{RGB}{31,119,180}
\definecolor{darkred}{RGB}{214,39,40}
\definecolor{mediumgray}{rgb}{0.5,0.5,0.5}
\definecolor{mediumteal}{rgb}{0,0.5,0.5}

\definecolor{ellisred}{rgb}{0.87,0.44,0.38} %
\definecolor{ellisgreen}{rgb}{0.69,0.90,0.52} %
\definecolor{elliscyan}{rgb}{0.29,0.77,0.74} %
\definecolor{ellisorange}{rgb}{0.89,0.55,0.28} %
\definecolor{ellisblue}{rgb}{0.41,0.61,0.86} %